\documentclass[10pt,twocolumn,letterpaper]{article}

\usepackage[pagenumbers]{cvpr} % To force page numbers, e.g. for an arXiv version

\usepackage{graphicx}
\usepackage{amsmath, mathtools}
\usepackage{bbm}
\usepackage{amssymb}
\usepackage{booktabs}
\usepackage{multirow}
\usepackage{makecell}   
\usepackage{pifont}
\newcommand{\cmark}{\ding{51}}%
\newcommand{\xmark}{\ding{55}}
\usepackage{algorithm, algpseudocode}
\newcommand{\keypoint}[1]{\vspace{0.1cm}\noindent\textbf{#1}\quad}
\usepackage[nolist]{acronym} 
\newcommand{\cut}[1]{}
\usepackage{wrapfig}
\usepackage[accsupp]{axessibility}

\usepackage{color, colortbl}
\definecolor{Gray}{gray}{0.9}
\definecolor{img-caption}{RGB}{150, 115, 166}
\definecolor{sub-caption}{RGB}{184, 84, 80}
\definecolor{img-retrieve}{RGB}{55, 0, 204}

\usepackage[pagebackref,breaklinks,colorlinks]{hyperref}

\usepackage[capitalize]{cleveref}
\crefname{section}{Sec.}{Secs.}
\Crefname{section}{Section}{Sections}
\Crefname{table}{Table}{Tables}
\crefname{table}{Tab.}{Tabs.}

\def\cvprPaperID{5032} % *** Enter the CVPR Paper ID here
\def\confName{CVPR}
\def\confYear{2023}

\begin{document}

\title{SceneTrilogy: On Human Scene-Sketch and its Complementarity \\ with Photo and Text\\[-0.4cm]}

\author{Pinaki Nath Chowdhury\textsuperscript{1,2} \hspace{.2cm}  Ayan Kumar Bhunia\textsuperscript{1} \hspace{.2cm} Aneeshan Sain\textsuperscript{1,2} \hspace{.2cm}  Subhadeep Koley\textsuperscript{1,2} \\
Tao Xiang\textsuperscript{1,2}\hspace{.3cm}  Yi-Zhe Song\textsuperscript{1,2} \\
\textsuperscript{1}SketchX, CVSSP, University of Surrey, United Kingdom.  \\
\textsuperscript{2}iFlyTek-Surrey Joint Research Centre on Artificial Intelligence.\\
{\tt\small \{p.chowdhury, a.bhunia, a.sain, s.koley, t.xiang, y.song\}@surrey.ac.uk} 
}
\vspace{-0.6cm}

\maketitle

%%%%%%%%% ABSTRACT
\begin{abstract}
In this paper, we extend scene understanding to include that of human sketch. The result is a complete trilogy of scene representation from three diverse and complementary {modalities} -- sketch, photo, and text. Instead of learning a rigid three-way embedding and be done with it, we focus on learning a flexible joint embedding that fully supports the ``optionality" that this complementarity brings. Our embedding supports optionality on two axes: (i) optionality across modalities -- use any combination of modalities as query for downstream tasks like retrieval, (ii) optionality across tasks -- simultaneously utilising the embedding for either discriminative (e.g., retrieval) or generative tasks (e.g., captioning). This provides flexibility to end-users by exploiting the best of each modality, therefore serving the very purpose behind our proposal of a trilogy in the first place. First, a combination of information-bottleneck and conditional invertible neural networks disentangle the modality-specific component from modality-agnostic in sketch, photo, and text. Second, the modality-agnostic instances from sketch, photo, and text are synergised using a modified cross-attention. Once learned, we show our embedding can accommodate a multi-facet of scene-related tasks, including those enabled for the first time by the inclusion of sketch, all without any task-specific modifications. Project Page: \url{http://www.pinakinathc.me/scenetrilogy}
\end{abstract}

%%%%%%%%% BODY TEXT
\vspace{-0.5cm}
\section{Introduction}
\vspace{-0.1cm}
\begin{figure}[t]%{l}{0.6\linewidth}
    % \vspace{-0.5cm}
    \centering
    \includegraphics[width=\linewidth]{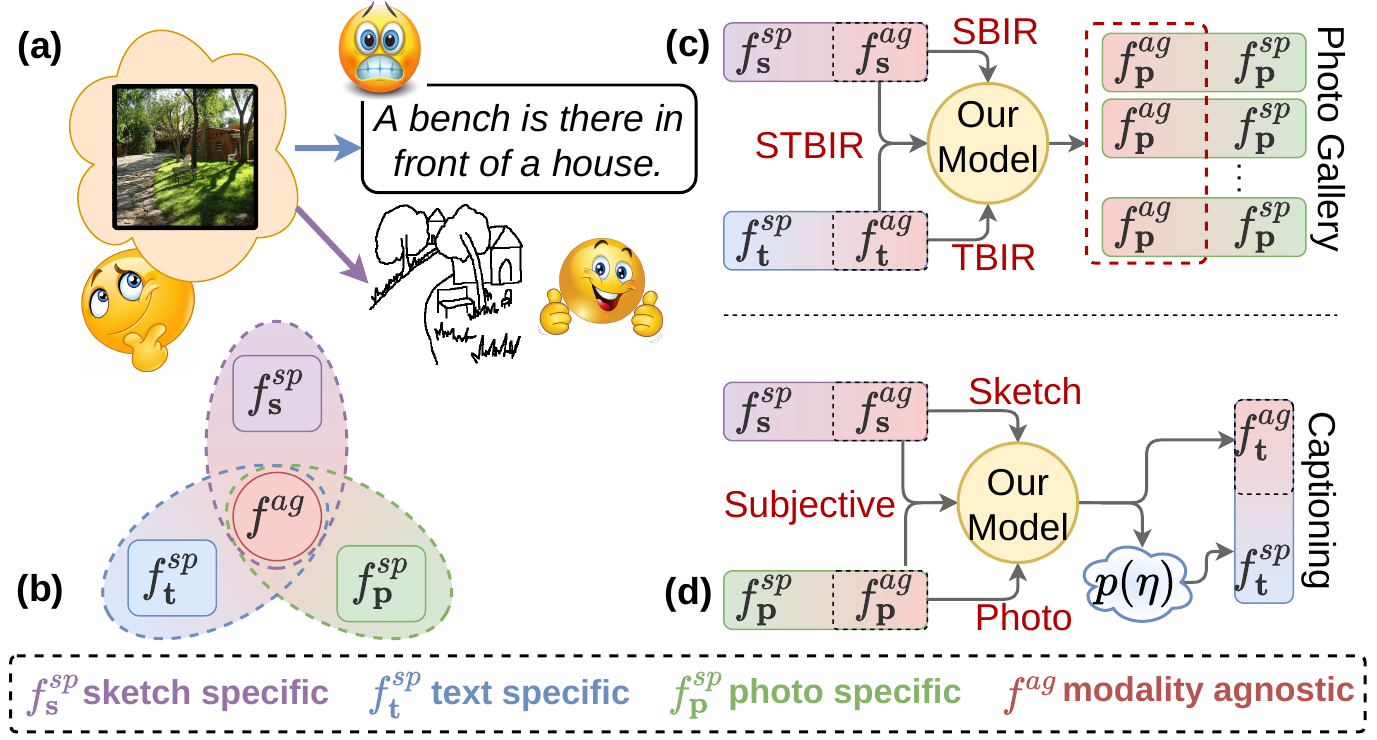}
    \vspace{-0.5cm}
    \caption{Some scenes are easy to describe via sketch; for others, text is better. We provide the option to sketch, write, or both (sketch+text). For ``optionality" across tasks, we disentangle sketch, text, and photo into a discriminative (e.g., retrieval) part $f^{ag}$ shared across modalities, and a generative (e.g., captioning) part specific to one modality ($f^{sp}_{\mathbf{s}}, f^{sp}_{\mathbf{t}} f^{sp}_{\mathbf{p}}$). This supports a multi-facet of scene-related tasks without task-specific modifications.}
    \label{fig:teaser}
    \vspace{-0.5cm}
\end{figure}

\label{sec:intro}
Scene understanding sits at the very core of computer vision. As object-level research matures \cite{imagenet, tuberlin-dataset}, an encouraging shift can be observed in recent years on scene-level tasks, e.g., scene recognition \cite{places-dataset}, scene captioning \cite{mscoco-dataset}, scene synthesis \cite{gao2020sketchyCOCO}, and scene retrieval \cite{castrejon2016cmplaces, liu2020scenesketcher}.

Scene research has generally progressed from that of single modality \cite{places-dataset, zhou2016semantic} to the very recent focus on multi-modality \cite{castrejon2016cmplaces, aytar2018crossmodal, fscoco}. The latter setting not only triggered a series of practical applications \cite{xu2015show, gao2020sketchyCOCO, liu2020scenesketcher, zou2018sketchyscene} but importantly helped to cast insights into scene understanding on a conceptual level (i.e., what is really being perceived by humans). To date, research on multi-modal scene understanding has mainly focused on two modalities -- text and photo \cite{liu2018discriminate, mahajan2020context, mahajan2020lnfmm}, via applications such as text-based scene retrieval (TBIR) \cite{luis2018textScene}, and scene captioning \cite{mahajan2020context, mahajan2020lnfmm, cornia2020m2net}.

This paper follows the said trend of multi-modal scene understanding and extend it to also include human scene-sketch. Sketch is identified because of its unique characteristics of being both expressive and subjective, evident in an abundance of object-level sketch research \cite{bhunia2022adaptive}, and very recently on scene-level \cite{fscoco}. To verify there is indeed useful complementarity that sketch can bring to multi-modal scene understanding, we first conducted two pilot studies (i) on expressivity, we compare text and sketch in terms of scene image retrieval, and (ii) on subjectivity, we test a novel task of subjective captioning where sketch or parts-of-speech \cite{deshpande2019pos} are used as guidance for image captioning. On (i), results show there is significant disagreement in terms of retrieval accuracy when one is used as query over the other, indicating there is complementary information between the two modalities. On (ii), sketch is shown to offer more subjectivity as a guiding signal than text, when quantified using common metrics such as {BELU-4 \cite{papineni2002bleu} and CIDEr \cite{vedantam2015cider}}.

To fully explore the complementarity of all three modalities, we desire a flexible joint embedding that best sustains ``optionality'' across \textit{modalities}, and also across \textit{tasks}. The former enables end-users to use any combination of modalities (e.g., only sketch, only text, or both sketch+text) as a query for downstream tasks; and the latter provides option of utilising the learned embedding for both discriminative (e.g., retrieval) and generative problems (e.g., captioning). 

This desired level of ``optionality" is however not achievable via naive three-way joint embeddings common in the literature \cite{castrejon2016cmplaces, aytar2018crossmodal, fscoco}. Instead, we advocate a three-way disentanglement (\cref{fig:teaser}(b)), where each of the three modalities is disentangled into their modality-specific component ($f^{sp}_{\mathbf{s}}$, $f^{sp}_{\mathbf{p}}$, $f^{sp}_{\mathbf{t}}$, for sketch, photo and text), and a shared modality-agnostic component ($f^{ag}$). The idea is that modality-specific will hold information specific to each modality (e.g., drawing style for sketch, texture for photo, and grammatical knowledge for text). It follows that filtering away modality-specific parts from \textit{each} of the three modalities gives a shared modality-agnostic part that carries shared abstract semantic across \textit{all} three modalities, (as shown in \cref{fig:teaser}(b)). How optionality is supported in such a disentangled space then becomes trivial (\cref{fig:teaser}(c),(d)). To achieve optionality across tasks, we simply use modality-agnostic information as the joint embedding to perform discriminative tasks (e.g., cross-modal retrieval), and for cross-modal generative tasks (e.g., captioning), we just combine modality-agnostic information (from source) with modality-specific (from target) to generate the target modality. Optionality across modality is a little harder, where we make use of a cross-attention \cite{set-attention} mechanism to capture the synergy across the modality-agnostic components.

{Benefiting from our optionality-enabled embedding, we can perform a multi-facet of tasks without any task-specific modifications: (i) \cref{fig:teaser} (c) show cross-modal discriminative tasks such as sketch-based image retrieval ({SBIR}) using ({$f^{ag}_{\mathbf{s}}$} $\leftrightarrow$ {$f^{ag}_{\mathbf{p}}$}), text-based image retrieval ({TBIR}) using ({$f^{ag}_{\mathbf{t}}$} $\leftrightarrow$ {$f^{ag}_{\mathbf{p}}$}), or sketch+text based image retrieval ({STBIR}) using ({$f^{ag}_{\mathbf{s}} + f^{ag}_{\mathbf{t}}$} $\leftrightarrow$ {$f^{ag}_{\mathbf{p}}$}). (ii) \cut{A by-product of disentangling modality-specific information is also a learned modality-specific prior $p(\eta)$ in \cref{fig:teaser}.} \cref{fig:teaser} (d) show cross-modal generative tasks such as image captioning ({photo} branch) using {$f^{ag}_{\mathbf{p}}$} $+$ {$f^{sp}_{\mathbf{t}}$} $\rightarrow$ $f_{\mathbf{t}}$ to generate textual descriptions $f_{\mathbf{t}}$. Similarly, for sketch captioning ({sketch} branch) we use {$f^{ag}_{\mathbf{s}}$} $+$ {$f^{sp}_{\mathbf{t}}$} $\rightarrow$ $f_{\mathbf{t}}$.
\cut{Sampling text-specific prior and combining with cross-modal image-agnostic information results in image captions (generative task). Similarly, combining sketch-agnostic information with text-specific prior helps generate sketch captions.} (iii) Last but not least, to demonstrate what the expressiveness of human sketch can bring to scene understanding, we introduce a novel task of subjective captioning where we guide image captioning using sketch as a signal ({subjective} branch) as {$f^{ag}_{\mathbf{p}}$} $+$ {$f^{ag}_{\mathbf{s}}$} $\rightarrow$ $f_{\mathbf{t}}$.

In summary, our contributions are: (i) We extend multi-modal scene understanding to include human scene-sketches, thereby completing a trilogy of scene representation from three diverse and complementary modalities. (ii) We provide optionality to end-users by learning a flexible joint embedding that supports: optionality across modalities and optionality across tasks. (iii) Using computationally efficient techniques like information bottleneck, conditionally invertible neural networks, and modified cross-attention mechanism, we model this flexible joint embedding. (iv) Once learned, our embedding accommodates a multi-facet of scene-related tasks like retrieval, captioning.

\section{Related Works}
\vspace{-0.2cm}
\keypoint{Sketch for Visual Understanding:}
Hand-drawn sketches enriched with human visual perception cues have facilitated several downstream visual understanding tasks. Apart from the widely explored SBIR \cite{liveSketch, bhunia2023sketch2saliency}, sketch has shown potential on object localisation \cite{chowdhury2023detect}, segmentation \cite{qi2022segmentation}, image/video synthesis \cite{koley2023picture}, representation learning \cite{sain2023exploiting}, 3D shape retrieval/modelling \cite{chowdhury20223Dsynthesis}, medical image analysis \cite{wang2022medical, kobayashi2023medical}, etc. \cite{xu2022survey}. Sketches are also useful in the creative industry like artistic image editing \cite{yang2020surgery} and animation \cite{xing2015autocomplete}. Unlike photos that are passively captured by a camera, sketches are drawn by humans that actively stimulate intelligence with pictionary-style drawing games \cite{pixelor}. While text has been widely used for human expression, in this paper, we show freehand sketches can provide complimentary or symbiotic information for visual understanding.

\keypoint{Sketch-Based Image Retrieval (SBIR):} SBIR retrieves a paired photo given a query sketch. Sketches offer visual description that commences the avenues of \emph{category-level} \cite{sain2023clip, yelamarthi2018sketch, doodle-to-search} or fine-grained \emph{instance-level} (FG-SBIR) \cite{bhunia2022worrying, bhunia2020sketch, bhunia2021semi} retrieval. SBIR typically employs deep triplet-ranking based siamese networks to learn a joint embedding space \cite{yu2016shoe}. Contemporary research emerged towards zero-shot SBIR \cite{doodle-to-search, sain2022sketch3t}, cross-domain translation \cite{kaiyue2017cross}, on-the-fly retrieval \cite{bhunia2020sketch}, semi-supervised \cite{bhunia2021semi}, self-supervised \cite{vector-raster}, meta-learning \cite{bhunia2022adaptive} etc. As research on object-level SBIR matured, focus shifted towards the more practical scene-level SBIR \cite{scene-designer} with GCN \cite{liu2020scenesketcher}, and optimal transport \cite{partially-does-it}. The onset of scene sketch datasets \cite{gao2020sketchyCOCO, zou2018sketchyscene, fscoco} revealed further insights into implicit human-sketching strategies \cite{fscoco}.

\keypoint{Text-Based Image Retrieval (TBIR):}
Learning image-text joint embedding space with ranking loss \cite{aviv2017tbir, plummer2017tbir, deep-visual-semantic} received considerable attention. Further improvements used mining hardest negative pairs for triplet loss \cite{faghri2017tbir}, cross-modal adaptive message passing \cite{camp}, probabilistic one-to-many representations \cite{chun2021pcme} etc. Despite text lacking visual cues, million-scale paired image-text datasets have made TBIR competitive due to power scaling laws \cite{song2021scaling}. This inspired large-scale methods like Oscar \cite{oscar}, and CLIP \cite{CLIP}. In this paper, we augment TBIR with sketches to provide the creativity and freedom of expression intrinsic to sketches.
 
\keypoint{Multi-Modality in Computer Vision:} 
Multi-modal learning (MML) aims at developing models that can extract, interpret, and reason on information from various modalities characterised by different statistical properties such as text, sketch, or text+sketch. Contemporary research studied MML in vision via image and text \cite{hong2021tbir}, image to scene graph \cite{guo2021scenegraph}, etc. \cite{xu2022MML}. 
MML faces challenges like cross-modal alignment \cite{jia2021scalingVL}, or efficiency over data \cite{touvron2021kd} and compute \cite{kim2021ViLT}. It is useful when data in one modality is inaccessible \cite{aytar2018crossmodal} for privacy or logistic reasons (e.g., hospital), but abundantly available in other modalities (photos in MS-COCO \cite{mscoco-dataset}). 
Often, some modalities are preferred over others for human-machine communication, like some concepts are easier to express in texts~\cite{lu2019ViLBERT}, while others prefer sketches~\cite{lin2020sketchbert} or both \cite{scene-designer} (\cref{fig:teaser}). In this paper, we learn cross-modal representation \cite{xue2021intermodality} that works using either one modality (text/sketch) or both.

\keypoint{Disentangled Representation for Multi-modality:}
Disentangling modality-agnostic from modality-specific residual factors is important for MML \cite{hsu2018disentanglement, learning-factorised-representations}. Modality-agnostic information is useful for cross-modal transfer like semantics-based retrieval and pattern recognition \cite{hsu2018disentanglement} but holds no meaning for tasks specific to one modality like image-style or speaker information \cite{suzuki2017disentanglement}. Disentanglement was explored where factors of variation are either known (e.g., facial poses~\cite{torontofacedataset}) and individually supervised \cite{learning-disentangled-factor}, partially known \cite{styleMeUp}, or unknown (e.g., drawing style \cite{styleMeUp}) and learned unsupervised using isotropic Gaussian prior \cite{rubenstein2018isotropic} or information-theoric regularisation \cite{infoGAN}. Our method aligns with the unknown setup where factors particular to sketch, text, and image are discovered unsupervised.

\keypoint{Image Captioning:} This has emerged from predicting syntactically correct descriptions \cite{zhang2022captioning,  stefanini2021captioningsurvey} to tackling data scarcity \cite{agarwal2019nocaps, laina2019captioningunsupervised}, and addressing user requirements \cite{park2017personalised, park2018personalised}. Predicted captions evolved from being factual in a neutral tone to (i) controllable using textual verbs \cite{chen2021captioningverbs}, part-of-speech tag \cite{deshpande2019pos}, or mouse trace \cite{pont2020mouse, meng2021mouse}, and (ii) personalised captioning \cite{zhang2020personalised, shuster2019personality} that learns user's active vocabulary, and writing style. Our method can (i) generate factual captions from images/sketches and (ii) extend controllable captioning paradigm by injecting saliency via sketch. %using \emph{sketch to inject saliency}.
% \vspace{-0.3cm}

\section{Pilot Study}
% \vspace{-0.1cm}
\subsection{Sketch vs. Text for Retrieval}
% \vspace{-0.4cm}

\begin{figure}[!htbp]
    \includegraphics[width=\linewidth]{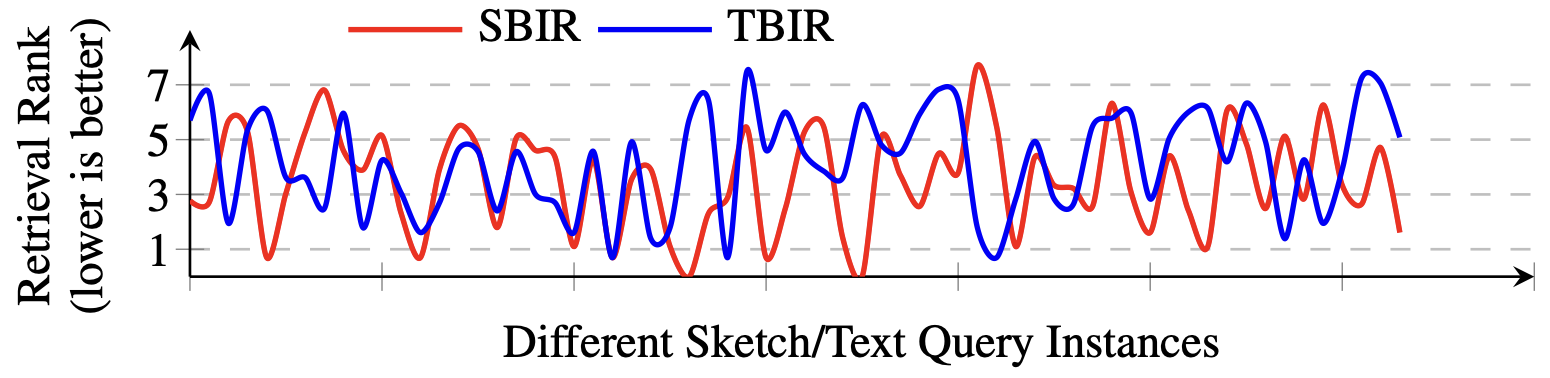}
    \vspace{-0.6cm}
	\caption{We compare SBIR \cite{yu2016shoe} vs. TBIR \cite{CLIP} on FS-COCO \cite{fscoco} where retrieval rank is plotted in \emph{log-scale} (see Supplemental for more details). While sketch is a better query for some instances (lower retrieval rank), for others text is better.}
    \label{fig:sbir-vs-tbir}
    \vspace{-0.3cm}
\end{figure}
% \begin{figure}[!htbp]
%     \centering
%     \includegraphics[width=0.9\linewidth]{cvpr23-figures/sbir-vs-tbir-v2.pdf}
%     \vspace{-0.3cm}
%     \caption{We compare SBIR vs. TBIR on FS-COCO \cite{fscoco} where retrieval rank is plotted in \emph{log-scale}. While sketch is a better query for some instances (lower retrieval rank), for others text is better.}
%     \label{fig:sbir-vs-tbir}
%     \vspace{-0.3cm}
% \end{figure}
\noindent Text can convey colour information, or object categories, but is cumbersome to describe fine-grained details, multiple objects, or complex shapes \cite{song2017textSketch}\footnote{Example: {Cross strap stud and buckle detail blonde leather upper leather insole chunky wooden sole $9$ cm heel.}}. While sketch can depict complex shapes, multiple objects, and spatial alignment \cite{fscoco}, not all objects are easy to draw (`donkey' vs. `horse').  Fig.~\ref{fig:sbir-vs-tbir} shows this trade-off between sketch vs. text for image retrieval. We find an optimal fusion between sketch and text to derive best of both modalities along with the ability to optionally use only sketch, only text, or both.

{
\setlength{\tabcolsep}{6.5pt}
\begin{table}[!h]
    \centering
    \footnotesize
    \caption{Comparing alternative guiding signal like POS (part-of-speech) \cite{deshpande2019pos}, Mouse Trace \cite{meng2021mouse}, and Freehand Sketches \cite{fscoco}. 
    % Sketch is competitive with Trace and better than POS. However, Sketch can represent additional information like artistic interpretation (caricature \cite{caricatureshop}).
    }
    \vspace{-0.3cm}
    \begin{tabular}{cccccccc}
        \toprule
        Signal & & B-1 & B-4 & M & R & C & S \\\hline
        \multirow{2}{*}{POS \cite{deshpande2019pos}} & w/o & 73.2 & 31.1 &  24.5 &  52.8 &  100.1 &  17.9 \\
          & w/ &  73.9 &  31.6 &  25.5 &  53.2 &  104.5 &  18.8 \\%\hdashline
         \rowcolor{Gray}
         $\Delta$ &  & \textbf{0.7} &  \textbf{0.5} &  \textbf{1.0} &  \textbf{0.4} &  \textbf{4.4} & \textbf{0.9} \\\hline
         \multirow{2}{*}{Trace \cite{meng2021mouse}} & w/o & 32.2 & 8.1 &  -- &  31.7 &  29.3 &  25.7 \\
          & w/ &  52.2 &  24.6 &  -- &  48.3 &  106.5 &  36.5 \\%\hdashline
         \rowcolor{Gray}
         $\Delta$ &  & \textbf{20} &  \textbf{16.5} &  \textbf{--} &  \textbf{16.6} &  \textbf{77.2} & \textbf{10.8} \\\hline
         \multirow{2}{*}{Sketch} & w/o & 74.7 & 31.8 &  24.7 &  53.8 &  105.5 &  18.8 \\
          & w/ &  81.3 &  42.7 &  30.1 &  61.6 &  121.6 &  23.5 \\%\hdashline
         \rowcolor{Gray}
         $\Delta$ &  & \textbf{6.6} &  \textbf{10.9} &  \textbf{5.4} &  \textbf{7.8} &  \textbf{16.1} & \textbf{4.7} \\\bottomrule
    \end{tabular} 
    \label{tab:subjective-captioning-schema}
    \vspace{-0.3cm}
\end{table}
}

\vspace{-0.1cm}
\subsection{Subjectivity for Captioning}\label{sec: pilot-captioning}
\vspace{-0.1cm}

\noindent Unlike traditional image captioning \cite{wang2017agcvae, mahajan2020context} that generates factual captions in neutral tone, subjective captioning adapts the predicted captions using a guiding signal that specifies priorities on what should be described \cite{stefanini2021captioningsurvey}. The signal is injected via feature concatenation \cite{deshpande2019pos}, or cross-attention mechanism \cite{meng2021mouse}. Applications of subjective captioning include medical report generation using disease tags to generate real style reports \cite{liu2021radiologycaptioning}, art descriptions \cite{artcaptioning}, and assistive technologies for the visually impaired \cite{alt-text, blindcaptioning}. In this paper, we advocate for sketch as a guiding signal to depict salient objects and express artistic interpretations \cite{caricatureshop}. We compare the performance (see supplementary for details) using guiding signals like POS (parts-of-speech)\cite{deshpande2019pos}, mouse trace \cite{meng2021mouse}, or freehand sketches \cite{fscoco}. Following \cite{meng2021mouse}, we inject the guiding signal into the image captioning pipeline via cross-attention mechanism. As evident from \Cref{tab:subjective-captioning-schema}, while sketch is competitive with mouse traces, it is a better signal than POS. However, unlike mouse trace, sketch can depict artistic interpretation \cite{artcaptioning} making it a more flexible and robust guiding signal than POS or mouse trace.

\section{Proposed Methodology}
\vspace{-0.1cm}
% A synergy between sketch/photo/text ---> 

% a) Sketch and/or Text for Retrieval

% b) Sketch-based subjective captioning. [alternatives --- why not only keywords,  ablation -- varying sketches -- varying text captions] 

\subsection{Preliminaries}
\vspace{-0.1cm}

\keypoint{Baseline for Fine-Grained Retrieval:}
Given a query-photo pair $(\mathbf{q}, \mathbf{p})$, existing methods encode \cite{yu2016shoe, li2018sketch-r2cnn, lin2020sketchbert, liu2020scenesketcher, vector-raster} the query $\mathbf{q} = \{\mathbf{s}, \mathbf{t}\}$ comprising sketch ($\mathbf{s}$) / text ($\mathbf{t}$) and photo ($\mathbf{p}$) as $f_{\mathbf{q}} = \mathcal{F}_{\mathbf{q}}(\mathbf{q}) \in \mathbb{R}^{D}$, and  $f_{\mathbf{p}} = \mathcal{F}_{\mathbf{p}}(\mathbf{p}) \in \mathbb{R}^{D}$ respectively. The network is trained via triplet loss with margin parameter $\mu > 0$ such that the cosine distance $\delta(\cdot)$ of query anchor $\mathbf{q}$ from a negative photo ($\mathbf{p}^{-}$) should increase while that from the positive photo ($\mathbf{p}^{+}$) should decrease as, $\mathcal{L}_{trip} = \max \{0, \mu + \delta(f_{\mathbf{q}}, f_{\mathbf{p}^{+}}) - \delta(f_{\mathbf{q}}, f_{\mathbf{p}^{-}}) \}$.

\keypoint{Baseline for Image Captioning:}
Image captioning consists of an image encoder \cite{xu2015show, liu2018discriminate}, $f_{\mathbf{p}} = \mathcal{F}_{\mathbf{p}}(\mathbf{p})$ followed by an autoregressive textual decoder ($\mathcal{F}_{C}$). Given the textual description comprises a sequence of words $\mathbf{t} = \{w_1, \dots, w_K\}$, we maximise the likelihood of a predicted word ($\hat{w}_{k}$) at each step ($k$), conditioned on $f_{\mathbf{p}}$ as, $\mathcal{L}_{C} = -\sum_{k=1}^{K} \log [\mathcal{F}_{C}(\hat{w}_{k} =w_{k} | f_{\mathbf{p}}, w_1, \dots, w_{k-1} )]$

\begin{figure}
    \centering
    \includegraphics[width=\linewidth]{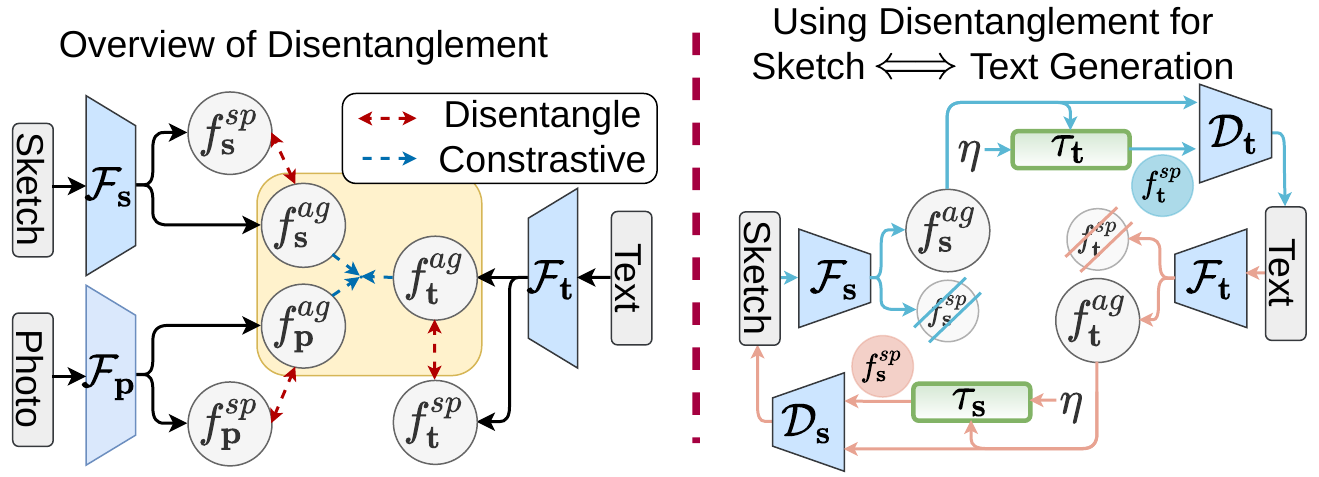}
    \vspace{-0.7cm}
    \caption{(Left): We disentangle modality-agnostic and modality-specific components from sketch, text, and photo. The modality-agnostic components are aligned using contrastive loss for cross-modal transfer. (Right): Modality-agnostic sketch ($f^{ag}_{\mathbf{s}}$) is used across modality to generate modality-specific text ($f^{sp}_{\mathbf{t}}$) using text-specific $\tau_{\mathbf{t}}$. Combining $f^{ag}_{\mathbf{s}}$ and $f^{ag}_{\mathbf{t}}$, we generate text from sketch.}
    \label{fig: overview}
    \vspace{-0.5cm}
\end{figure}

\vspace{-0.1cm}
\subsection{Overview}
\vspace{-0.2cm}

\noindent We aim to disentangle the feature representations from sketch, text, and photo modalities into a \emph{modality-agnostic} and \emph{modality-specific} component. While the modality-\textit{agnostic} component holds semantic information to support cross-modal transfer, the modality-\textit{specific} one holds information necessary during self-reconstruction; however, it lacks meaning in other modalities (e.g., grammatical knowledge in text). Achieving feature disentanglement across scene sketches, texts, and photos enables a multitude of downstream tasks like (i) \emph{SBIR} -- modality-agnostic sketch and photo features, (ii) \emph{TBIR} -- modality-agnostic text and photo, (iii) \emph{Sketch+Text-Based Image Retrieval} -- modality-agnostic sketch, text, and photo, (iv) \emph{Image Captioning} -- using the modality-agnostic photo to compute modality-specific text features, (v) \emph{Sketch Captioning} -- modality-agnostic sketch to compute modality-specific text, and (vi) \emph{Subjective Captioning} -- using modality-agnostic photo and sketch, to compute modality-specific text.
% and computing

\vspace{-0.1cm}
\subsection{Disentangling Modality Agnostic and Specific}\label{sec: two-modality}
\vspace{-0.2cm}

%two-modality
\noindent While our disentangling method can be generalised to any number of modalities, for simplicity, we first show for $M=2$ modalities and later extend to $M \geq 3$. Consider a simple bimodal setup of sketch ($\mathbf{s} \in \mathbb{R}^{H \times W \times 3}$) and text ($\mathbf{t} \in \mathbb{R}^{N \times E}$). Our goal is to split the feature representation $f_{\mathbf{s}} = \mathcal{F}_{\mathbf{s}}(\mathbf{s}) \in \mathbb{R}^{512}$ and $f_{\mathbf{t}} = \mathcal{F}_{\mathbf{t}}(\mathbf{t}) \in \mathbb{R}^{512}$ into a modality-\textbf{ag}nostic and a modality-\textbf{sp}ecific component as $f_{\mathbf{s}} = [f^{ag}_{\mathbf{s}}, f^{sp}_{\mathbf{s}}]$, and $f_{\mathbf{t}} = [f^{ag}_{\mathbf{t}}, f^{sp}_{\mathbf{t}}]$ respectively, where $f^{ag} \in \mathbb{R}^{480}$ and $f^{sp} \in \mathbb{R}^{32}$. Existing methods \cite{spurr2018crossmodal, styleMeUp} disentangle feature representations via (i) self reconstruction as $\hat{\mathbf{s}} = \mathcal{D}_{\mathbf{s}}([f^{ag}_{\mathbf{s}}, f^{sp}_{\mathbf{s}}])$ and $\hat{\mathbf{t}} = \mathcal{D}_{\mathbf{t}}([f^{ag}_{\mathbf{t}}, f^{sp}_{\mathbf{t}}])$ coupled with (ii) cross-modal translation $\hat{\mathbf{s}} = \mathcal{D}_{\mathbf{s}}([f^{ag}_{\mathbf{t}}, f^{sp}_{\mathbf{s}}])$ and $\hat{\mathbf{t}} = \mathcal{D}_{\mathbf{t}}([f^{ag}_{\mathbf{s}}, f^{sp}_{\mathbf{t}}])$. However, using cross-modal translation with latent feature exchange across modalities is a cumbersome process that explodes with $\mathbb{P}^M_{2}$ permutations for $M$ modalities, e.g., $M=3$ has $\mathbb{P}^{3}_{2}=6$ cross-modal translations. Adding multiple cross-modal translation losses makes optimisation difficult and computationally expensive. We break this compute barrier with linear ($\mathcal{O}(M)$) complexity using an \emph{information bottleneck reinterpretation} of modality-agnostic and modality-specific disentanglement. In particular, we \emph{maximise} the  \emph{mutual information} $\mathcal{I}(f^{ag}_{\mathbf{s}}, f^{ag}_{\mathbf{t}})$ amongst modality-agnostic components, while \emph{minimising} the \emph{same} between modality-agnostic and modality-specific components $\mathcal{I}(f^{ag}_{\mathbf{s}}, f^{sp}_{\mathbf{s}})$, and $\mathcal{I}(f^{ag}_{\mathbf{t}}, f^{sp}_{\mathbf{t}})$, 
where $\mathcal{I}$ ($\cdot$,$\cdot$) denotes mutual information between two entities.
Hence, unlike the previous $\mathbb{P}^M_2$ permutations, \cref{eq: loss-disentanglement} has one agnostic $\mathcal{I}(f^{ag}_{\mathbf{s}}, f^{ag}_{\mathbf{t}})$, and $M$ specific $\mathcal{I}(f^{ag}_{\mathbf{k}}, f^{sp}_{\mathbf{k}})$ losses. Formally, using a Langrange multiplier hyperparameter $\beta$ we have our loss objective as,
\vspace{-0.3cm}
\begin{equation}\label{eq: loss-disentanglement}
    \mathcal{L}_\mathcal{I} = - \overbrace{\mathcal{I}(f^{ag}_{\mathbf{s}}, f^{ag}_{\mathbf{t}})}^{agnostic} + \beta \overbrace{\sum_{\mathbf{k} \in \{\mathbf{s}, \mathbf{t}\}} \mathcal{I}(f^{ag}_{\mathbf{k}}, f^{sp}_{\mathbf{k}})}^{specific}
\end{equation}
\vspace{-0.6cm}

\begin{figure}
    \centering
    \includegraphics[width=\linewidth]{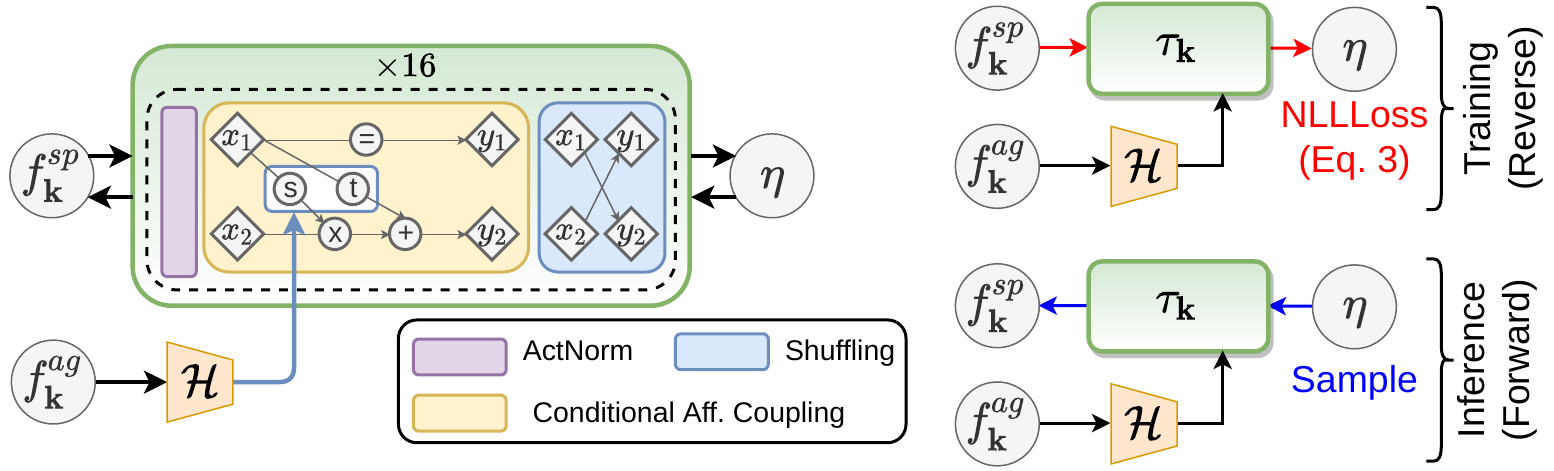}
    \vspace{-0.7cm}
    \caption{Unlike typical neural networks that are unidirectional, data in conditional invertible neural networks ($\tau_{\mathbf{k}}$) can flow either (i) from modality-specific $f^{sp}_{\mathbf{k}}$ to a uniform distribution $\eta$ by conditioning on modality-agnostic $f^{ag}_{\mathbf{k}}$ \textbf{during training}, or (ii) from a sampled $\eta$ in uniform distribution to the modality-specific $f^{sp}_{\mathbf{k}}$ by conditioning on modality-agnostic $f^{ag}_{\mathbf{k}}$ \textbf{during inference}. The conditioning vector $f^{ag}_{\mathbf{k}}$ is injected into the conditional affinity coupling layers \cite{nice} of $\tau_{\mathbf{k}}$ using any arbitrary network $\mathcal{H}$.}
    \label{fig:cINN-schematic}
    \vspace{-0.5cm}
\end{figure}

\begin{figure*}
    \centering
    \includegraphics[width=\linewidth]{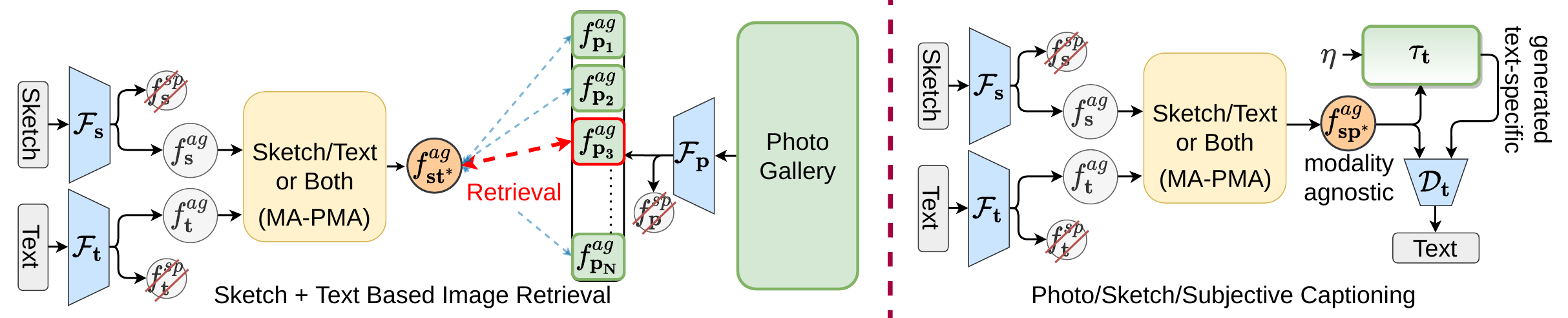}
    \vspace{-0.7cm}
    \caption{(Left): The modality-agnostic from sketch, or text, or both are used to retrieve from a gallery of photos. This enables a multitude of retrieval tasks like SBIR, TBIR, and STBIR. (Right): The modality-agnostic from photo, or sketch, or both are used to generate the text-specific component. Combining the modality-agnostic and inferred text-specific (via $\tau_{\mathbf{t}}$) enables image, or sketch, or subjective captioning.}
    \label{fig: optinal-sketch-text}
    \vspace{-0.5cm}
\end{figure*}

% (Conditional Aff. Coupling)
\keypoint{Minimise $\mathcal{I}(f^{ag}_{\mathbf{k}}, f^{sp}_{\mathbf{k}})$:} We minimise the mutual information between modality-agnostic and modality-specific components using a conditional invertible \cite{} neural network $\tau_{\mathbf{k}}$. Unlike typical unidirectional neural networks $\mathcal{F}: x \rightarrow y$, a conditional invertible neural network employs a sequence of bijective mapping operations like activation normalization (ActNorm) \cite{glow}, Conditional Affine Coupling~\cite{nice}, and shuffling \cite{glow} to obtain $\tau_{\mathbf{k}}: x \leftrightarrow y$. During the \emph{forward pass} {(inference)}, we sample $\eta \in \mathbb{R}^{32}$ from a uniform prior distribution $\mathbbm{p}(\eta)$ to predict the modality-specific $f^{sp}_{\mathbf{k}} \in \mathbb{R}^{32}$ by conditioning on $f^{ag}_{\mathbf{k}}$ as, $f^{sp}_\mathbf{k} = \tau_{\mathbf{k}} ( \eta \ | \ f^{ag}_{\mathbf{k}} )$. {In other words, during inference, we predict the modality-specific component of target from the modality-agnostic one of input using $\tau_{\mathbf{k}}$. The target modality is then generated by combining the input-agnostic and target-specific components.} The conditioning modality-agnostic vector $f^{ag}_{\mathbf{k}}$ is injected into the intermediate conditional affine coupling layers $\mathcal{C}: x \leftrightarrow y$ as: $[x_1, x_2] = \texttt{split} (x)$, and $y = \texttt{concat}[x_1, s_{\theta}  ([x_1; h]) \odot x_2 + t_{\theta}([x_1; h]) ]$, where, $h = \mathcal{H}(f^{ag}_{\mathbf{k}})$. A simple feed-forward neural network implements $s_{\theta}$, $t_{\theta}$, and $\mathcal{H}$. We learn $\tau_{\mathbf{k}}$ in the \emph{reverse pass} {(training)} via negative log-likelihood (NLL Loss in \cref{fig:cINN-schematic}) of $\tau^{-1}_{\mathbf{k}}( f^{sp}_{\mathbf{k}} \ | \ f^{ag}_{\mathbf{k}} )$ to predict a uniform distribution $\mathbbm{p}(\eta)$,
\begin{equation}\label{eq: cINN-loss}
    \mathbbm{p}(\eta) = \mathbbm{p}(\tau^{-1}_{\mathbf{k}}( f^{sp}_{\mathbf{k}} \ | \ f^{ag}_{\mathbf{k}} )) \ | \mathrm{det} J_{\tau^{-1}_{\mathbf{k}}}( f^{sp}_{\mathbf{k}} \ | \ f^{ag}_{\mathbf{k}} ) |
\end{equation}
{We show how learning $\tau_{\mathbf{k}}$ in \cref{eq: cINN-loss} minimises $\mathcal{I}(f^{ag}_{\mathbf{k}}, f^{sp}_{\mathbf{k}})$.} $\mathcal{I}(f^{ag}_{\mathbf{k}}, f^{sp}_{\mathbf{k}}) = \int_{f^{sp}_{\mathbf{k}}} \ \mathbbm{p}(f^{sp}_{\mathbf{k}} | f^{ag}_{\mathbf{k}}) \ \log \ {\mathbbm{p}(f^{sp}_{\mathbf{k}} | f^{ag}_{\mathbf{k}})}/{\mathbbm{p}(f^{sp}_{\mathbf{k}})}$. Approximating modality-specific prior $\mathbbm{p}(f^{sp}_{\mathbf{k}})$ with variational distribution $\mathbbm{q}(f^{sp}_{\mathbf{k}})$ gives the upper-bound, minimising which reduces the KL-divergence between $\mathbbm{p}(f^{sp}_{\mathbf{k}} | f^{ag}_{\mathbf{k}})$ and $\mathbbm{q}(f^{sp}_{\mathbf{k}})$ i.e., it encourages the disentanglement $\mathbbm{p}(f^{ag}_{\mathbf{k}}, f^{sp}_{\mathbf{k}})$ $\approx$ $\mathbbm{p}(f^{ag}_{\mathbf{k}}) \cdot \mathbbm{p}(f^{sp}_{\mathbf{k}})$. The prior $\mathbbm{q}(f^{sp}_{\mathbf{k}})$ is solved using $\tau_{\mathbf{k}}$ to enforce disentanglement between modality-agnostic and modality-specific components, like that in \cref{eq: cINN-loss}, as the sum of \emph{negative-loglikelihood} (NLL-Loss in \cref{fig:cINN-schematic}) and \emph{log-determinant} (see supplementary for proof),
\vspace{-0.2cm}
\begin{equation} 
\begin{split}
    \mathcal{L}_{\tau_{\mathbf{k}}} = &  - \mathbb{E}_{f^{sp}_{\mathbf{k}}} \{ \log \mathbbm{q}(\tau^{-1}_{\mathbf{k}}(f^{sp}_{\mathbf{k}} \ | \ f^{ag}_{\mathbf{k}}))\\
     & \hspace{5em} + \log|\mathrm{det} J_{\tau^{-1}_{\mathbf{k}}}(f^{sp}_{\mathbf{k}} \ | \ f^{ag}_{\mathbf{k}})| \}
\end{split}
\end{equation}
\vspace{-0.3cm}

\keypoint{Maximise $\mathcal{I}(f^{ag}_{\mathbf{s}}, f^{ag}_{\mathbf{t}})$:} Here we show how minimising a {constrastive based retrieval loss} \cite{cpc2018} between the modality-agnostic components of sketch and text will maximise their mutual information. We define contrastive loss matching modality-agnostic components of sketch and text as,
\begin{equation}\label{eq: contrastive}
    \mathcal{L}_{cl}^{\mathbf{s, t}} = - \mathbb{E}_{f^{sp}_{\mathbf{s}}} \Bigg[ \log \frac{ \omega (f^{ag}_{\mathbf{s}}, f^{ag}_{\mathbf{t}^{+}}) }{ \omega (f^{ag}_{\mathbf{s}}, f^{ag}_{\mathbf{t}^{+}}) + \sum^{N-1}_{f^{ag}_{\mathbf{t}^{-}}} \ \omega (f^{ag}_{\mathbf{s}}, f^{ag}_{\mathbf{t}^{-}}) } \Bigg]
\end{equation}
where, $\omega = \exp ( x^{T} \ \mathbf{W} \ y )$. For each modality-agnostic $f^{ag}_{\mathbf{s}}$ we sample a positive $f^{ag}_{\mathbf{t}^{+}}$ and $(N-1)$ negative $f^{ag}_{\mathbf{t}^{-}}$ pairs. The contrastive loss in \cref{eq: contrastive} is expressed as mutual information between $f^{ag}_{\mathbf{s}}$ and $f^{ag}_{\mathbf{t}}$ as, $\mathcal{L}^{\mathbf{s,t}}_{cl} \geq -\mathcal{I}(f^{ag}_{\mathbf{s}}, f^{ag}_{\mathbf{t}}) + \log(N)$. Hence, to maximise the mutual information between modality-agnostic $f^{ag}_{\mathbf{s}}$ and $f^{ag}_{\mathbf{t}}$, we can maximise the tractable lower bound $\log(N) - \mathcal{L}^{\mathbf{s, t}}_{cl}$.

\keypoint{Total Loss for Bimodal Setup:} The resulting loss ($\mathcal{L}_{tot}$) for bimodal (sketch and text) setup comprise three loss objectives (i) \emph{self reconstruction} loss $\mathcal{L}_{rec}$, (ii) \emph{contrastive loss} between two modality-agnostic terms $\mathcal{L}_{cl}^{\mathbf{s, t}}$, and (iii) \emph{disentanglement} between modality-agnostic and modality-specific components in each modality ($\mathbf{k}$) ($\mathcal{L}_{\tau_{\mathbf{k}}}$), as
\vspace{-0.2cm}
\begin{equation}\label{eq: total-loss-2}
\begin{split}
    \mathcal{L}_{rec} & = || \mathbf{s} - \mathcal{D}_{\mathbf{s}}(\mathcal{F}_{\mathbf{s}}(\mathbf{s})) ||_{2} + || \mathbf{t} - \mathcal{D}_{\mathbf{t}}(\mathcal{F}_{\mathbf{t}}(\mathbf{t})) ||_{2} \\
    &  \mathcal{L}_{tot} = \mathcal{L}_{rec} + \mathcal{L}_{cl}^{\mathbf{s, t}} + \beta [\mathcal{L}_{\tau_{\mathbf{s}}} + \mathcal{L}_{\tau_{\mathbf{t}}}]
\end{split}
\end{equation}

\keypoint{Extending to Three/More Modalities:} 
Here we extend our bimodal setup in \cref{sec: two-modality} to three or more modalities. 
(i) We compute the self-reconstruction loss for three modalities as $\mathcal{L}_{rec} = \sum_{\mathbf{k} \in \{ \mathbf{s}, \mathbf{t}, \mathbf{p} \}} || \mathbf{k} - \mathcal{D}_{\mathbf{k}}(\mathcal{F}_{\mathbf{k}}(\mathbf{k})) ||_{2}$. (ii) we minimise the mutual information between modality-agnostic and modality-specific components for sketch, text, and photo as, $\mathcal{L}_{\tau} = \mathcal{L}_{\tau_{\mathbf{s}}} + \mathcal{L}_{\tau_{\mathbf{t}}} + \mathcal{L}_{\tau_{\mathbf{p}}}$. (iii) However, our contrastive loss term $\mathcal{L}_{cl}$ that maximises the mutual information among modality-agnostic components can only compare two modalities. We can extend this naively to a three-modality setup as $\mathcal{L}_{cl}^{tot} = \mathcal{L}^{\mathbf{s, t}}_{cl} + \mathcal{L}^{\mathbf{s, p}}_{cl} + \mathcal{L}^{\mathbf{t, p}}_{cl}$.

{Extending to three or more modalities, however, we notice our contrastive loss in \cref{eq: contrastive} is defined for only bimodal setup ($\mathcal{L}_{cl}^{\mathbf{s,t}}$, or $\mathcal{L}_{cl}^{\mathbf{s,p}}$, or $\mathcal{L}_{cl}^{\mathbf{t,p}}$). For example, given three modalities $\mathcal{S}_{M} = \{m_1, m_2, m_3\}$, comparing only ($m_1, m_2$) ignores $m_3$. This highlights a key limitation: it fails when we have a query in both ($m_1, m_3$) to retrieve $m_2$ (e.g., sketch+text for image retrieval). Now the research question boils down to -- how can we model a function $\mathcal{G}(\cdot)$ such that it can model either $m_1$, or $m_3$, or both ($m_1, m_3$) to retrieve $m_2$. To design $\mathcal{G}$, using naive addition as $\mathcal{G}(m_1, m_3) = m_1 + m_3$ does not handle overlapping or conflicting information\footnote{When signals ($m_1, m_3$) are similar or complementary $\mathcal{G}$ should strengthen decision; when signals conflict $\mathcal{G}$ should filter unreliable ones.} in $m_1$ and $m_3$ \cite{learn-to-combine}. While, concatenation $\mathcal{G}(m_1, m_3) = \texttt{concat}[m_1, m_3]$ computes interaction between ($m_1, m_3$), it forces to provide both $m_1$ and $m_3$ during inference; thereby failing to model either $m_1$, or $m_3$, or both ($m_1, m_3$).}

\vspace{-0.1cm}
\subsection{Modelling Optional Sketch or Text}\label{sec: set-attention}
\vspace{-0.2cm}

\noindent We propose a simple approach to design $\mathcal{G}$ that optionally models either $m_1$, or $m_3$, or both ($m_1, m_3$), and handles overlapping or conflicting information. Our proposed $\mathcal{G}$ comprises a multihead cross-attention module $\texttt{MH}(\cdot)$ followed by an attention-based pooling $\texttt{PMA}(\cdot)$ as, $f_{M} = \texttt{PMA}(H_{M})$; where $H_{M} = \texttt{MH}(\mathcal{S}_{M})$, and $\mathcal{S}_{M} = \{m_1, m_3 \}$.

Our $\texttt{MH}(\cdot)$ is order-invariant and independent of the number ($M$) of input modalities defined as $\texttt{MH}(X) = \sigma(X X^{T}) X$; where $\sigma$ is scaled-softmax, $X^{T}$ is transpose of $X$, and $X \in \mathbb{R}^{M \times 480}$ is a list of modality-agnostic components $m_1$, or $m_3$ with $\mathbb{R}^{1 \times 480}$, or $(m_1, m_3) \in \mathbb{R}^{2 \times 480}$ in query. The cross-attention in $\texttt{MH}(\cdot)$ interacts across query modalities to compute mutually agreeing information between ($m_1, m_3$) as, $H_{M} \in \mathbb{R}^{2 \times 480}$. Next, we use an order-invariant attention-based pooling $\texttt{PMA}: \mathbb{R}^{2 \times 480} \rightarrow \mathbb{R}^{1 \times 480}$ with a learned seed vector $\mathcal{P} \in \mathbb{R}^{1 \times 480}$ to aggregate mutually agreeing $H_{M}$ as, $f_{M} = \texttt{PMA}(H_{M}) = \sigma(\mathcal{P} H_{M}^{T}) H_{M}$.
Hence, using our proposed fusion module $\mathcal{G}$, we adapt our contrastive loss defined for only a pair of modality-agnostic components in \cref{eq: contrastive} as $\mathcal{L}_{cl}^{tot} = \mathcal{L}^{\mathbf{s, t}}_{cl} + \mathcal{L}^{\mathbf{s, p}}_{cl} + \mathcal{L}^{\mathbf{t, p}}_{cl}$ to jointly model sketch--text--photo (or more) modality-agnostic as: $\mathcal{L}_{cls}^{tot} = \mathcal{L}_{cl}(\mathcal{G}(f^{ag}_{\mathbf{s}}, f^{ag}_{\mathbf{t}}), f^{ag}_{\mathbf{p}}) + \mathcal{L}_{cl}(\mathcal{G}(f^{ag}_{\mathbf{s}}, f^{ag}_{\mathbf{p}}), f^{ag}_{\mathbf{t}}) + \mathcal{L}_{cl}(\mathcal{G}(f^{ag}_{\mathbf{p}}, f^{ag}_{\mathbf{t}}), f^{ag}_{\mathbf{s}}) $. For a generalised solution involving more than three modalities ($M>3$), see supplementary.

\keypoint{Inference Data Flow:} We describe the inference data flow in \cref{fig: optinal-sketch-text}. For retrieval tasks, we first compute the modality-agnostic component of query sketch and text ($f^{ag}_{\mathbf{s}},f^{ag}_{\mathbf{t}}$), and a gallery of photos $\{ f^{ag}_{\mathbf{p_1}}, f^{ag}_{\mathbf{p_2}}, \dots, f^{ag}_{\mathbf{p_N}} \}$. Next, a combined representation for either only sketch ($f^{ag}_{\mathbf{s}}$), or only text ($f^{ag}_{\mathbf{t}}$), or both ($f^{ag}_{\mathbf{s}}, f^{ag}_{\mathbf{t}}$) is computed using multihead cross attention $\texttt{MH}(\cdot)$ followed by attention-based pooling $\texttt{PMA}(\cdot)$ defined in \cref{sec: set-attention} to get $f^{ag}_{\mathbf{st}^{*}}$. Finally, we find the minimum distance between the combined $f^{ag}_{\mathbf{st}^{*}}$ and modality-agnostic component of photo $f^{ag}_{\mathbf{p}_{i}}$ as $\omega(f^{ag}_{\mathbf{st}^{*}}, f^{ag}_{\mathbf{p_i}})$ defined in \cref{eq: contrastive}. For captioning, we additionally use the text-specific conditional invertible neural network $\tau_{\mathbf{t}}$ to generate the target modality-specific text (e.g., grammatical structure etc.) from input modality-agnostic comprising of only photo ($f^{ag}_{\mathbf{p}}$) for image captioning, only sketch ($f^{ag}_{\mathbf{s}}$) for sketch captioning, or both photo and sketch ($f^{ag}_{\mathbf{p}}, f^{ag}_{\mathbf{s}}$) for subjective captioning (i.e., generate image captions by conditioning on the input sketch).

% \vspace{-0.05cm}
\section{Experiments}
\vspace{-0.2cm}
\keypoint{Datasets:} We use two scene sketch datasets with fine-grained alignment among sketch, text, and photo: (i) SketchyCOCO \cite{gao2020sketchyCOCO} contains $14,081$ scene sketch-photo pairs. The photos are taken from MS-COCO \cite{mscoco-dataset} comprising $164K$ photos with paired texts. However, most sketches in SketchyCOCO \cite{gao2020sketchyCOCO} contain less than one foreground instance. Following \cite{liu2020scenesketcher}, we filter SketchyCOCO with one foreground instance to get $1015/210$ train/test scene sketches. (ii) Unlike SketchyCOCO \cite{gao2020sketchyCOCO}, where the scene sketches are synthetically generated, FS-COCO \cite{fscoco} includes $7000/3000$ train/test human-drawn scene sketches with a paired textual description of sketches.

\keypoint{Implementation Details:} Our model is implemented in PyTorch using 11GB Nvidia RTX 2080-Super GPU. First, we pre-train the image encoder and text decoder for image captioning using $82,783$ photo-text pairs (excluding the photos common in SketchyCOCO and FS-COCO) for $15$ epochs. Next, we fine-tune on either SketchyCOCO \cite{gao2020sketchyCOCO}, or FS-COCO \cite{fscoco} for $200$ epochs using Adam optimiser with learning rate $1e-4$, and batch size $64$. Our photo ($\mathcal{F}_{\mathbf{p}}$) and sketch ($\mathcal{F}_{\mathbf{s}}$) encoders use ImageNet pretrained VGG-16 \cite{vgg-16}. For simplicity, we encode text using a bidirectional GRU unit with $512$ hidden units. Our text decoder \cite{deep-visual-semantic} is a single-layer autoregressive LSTM decoder that predicts a probability distribution over a fixed vocabulary ($10,010$ words) at every time step. For the image/sketch decoder, we use two separate GAN \cite{self-atten-gan} networks that synthesise sketch/image of size $64\times 64$, respectively. For brevity, we avoid realistic sketch/image generation due to the challenging scene complexity \cite{fscoco}. Hence we do not use a discriminator module for high-quality, sharp reconstruction \cite{stackgan}. Finally, our conditionally invertible neural network comprises $16$ alternating affine coupling \cite{realNVP}, activation normalisation \cite{glow}, and switch layers \cite{realNVP}.

\keypoint{Evaluation Metric:} In line with FG-SBIR research, we use Acc.@q \cite{sain2020crossmodal} defined as the percentage of sketches having a true matched photo in the top-q list. For sketch/image/subjective captioning, we use standard metrics BELU (B) 1-4 \cite{papineni2002bleu}, CIDEr (C) \cite{vedantam2015cider}, ROUGE (R) \cite{lin2004rouge}, METEOR (M) \cite{denkowski2014meteor}, and SPICE \cite{anderson2016spice}. Following \cite{wang2017agcvae}, we generate $100$ candidate captions and employ consensus re-ranking using CIDEr to select the best candidate caption.

\keypoint{Competitors:} We compare against (i) existing state-of-the-art methods that align two modalities (S2): For SBIR, \textbf{Triplet-SN} \cite{yu2016shoe} employs Sketch-A-Net \cite{sketch-a-net} backbone trained using triplet loss. \textbf{HOLEF} \cite{deep-spatial-semantic} adds spatial attention with a higher-order ranking loss. \textbf{SketchyS} \cite{zou2018sketchyscene} replaces Sketch-A-Net in \emph{Triplet-SN} with VGG-16 \cite{vgg-16} and an auxiliary category-level cross-entropy. \textbf{SceneS} \cite{liu2020scenesketcher} uses GCN \cite{gcn} to model scene sketch layout information. For TBIR, \textbf{CLIP} \cite{CLIP} is trained with text using transformer \cite{transformer} and photo using vision transformer \cite{VIT} on $400$ million text-photo pairs. \textbf{CLIP-LN} fine-tunes \emph{CLIP} by training only layer normalisation parameters \cite{layer-norm} with learning rate $0.00001$. For image/sketch captioning, \textbf{SAT} \cite{xu2015show} is one of the simplest but seminal works using a CNN-LSTM encoder-decoder approach similar to ours. \textbf{GMM-CVAE} \cite{wang2017agcvae} employs a conditional variational autoencoder with a Gaussian mixture model. \textbf{LNFMM} \cite{mahajan2020lnfmm} is similar to ours that splits information into modality-agnostic and modality-specific components using conditional invertible neural network, \textbf{ClipCap} \cite{clipcap} employs CLIP \cite{CLIP} for image encoding followed by GPT-2 \cite{gpt2} for text decoding. A learned mapping module translates CLIP embeddings to GPT-2. (ii) We compare against methods that align $3$ modalities (S3): For STBIR, \textbf{QST} \cite{song2017textSketch} extends triplet loss in \emph{Triplet-SN} to quadruplet loss that combines sketch and text for image retrieval. \textbf{SCM} uses element-wise addition to combine sketch and text from ResNet-18 \cite{resnet} with weight sharing across sketch, text, and photo from \emph{ResBlock4} onwards \cite{aytar2018crossmodal}. (iii) We design baselines (B): For STBIR, \textbf{CrossAtt} employs cross-attention \cite{transformer} to combine sketch and text. For subjective captioning, \textbf{MulCap} combines sketch ($f^{ag}_{\mathbf{s}}$) and photo ($f^{ag}_{\mathbf{p}}$) via element-wise multiplication as in \cite{changpinyo2019captioning}. \textbf{CrossCap} optionally fuse photo, sketch, or both using cross-attention. \textbf{CatCap} use feature concatenation \cite{pont2020mouse} of guiding sketch ($f^{ag}_{\mathbf{s}}$) signal with photo ($f^{ag}_{\mathbf{p}}$) to generate captions.

\vspace{-0.1cm}
\subsection{Combining Sketch and Text for Image Retrieval}
\vspace{-0.2cm}
\noindent \cref{fig:sbir-vs-tbir} shows that for some instances, sketch is a better query, whereas text is better for others. Hence, to achieve \emph{best of both modalities}, we examine the complimentary nature by combining sketch and text for image retrieval. \Cref{tab:stbir} shows (i) \emph{SCM} gives the lowest performance due to naive element-wise addition of potentially overlapping and conflicting information \cite{learn-to-combine} from sketch and text. (ii) \emph{QST} improves slightly upon \emph{SCM} by replacing naive element-wise addition with a weighted summation ($0.8$ for sketch modality). (iii) \emph{CrossAtt} outperforms all baselines by using a cross-attention between sketch and text to resolve overlapping/conflict information \cite{learn-to-combine}. (iv) Our proposed method gives the highest performance due to cross-attention that model sketch-text interaction and disentanglement to drive out modality-specific information for cross-modal retrieval.

{
\begin{table}[!t]
    \centering
    \footnotesize
    \caption{Quantitative results combining sketch and text for image retrieval (FG-STBIR) on two scene sketch datasets \cite{gao2020sketchyCOCO, fscoco}.}
    \vspace{-0.3cm}
    \begin{tabular}{clcc|cc}
        \toprule
        \multicolumn{2}{c}{\multirow{2}{*}{Method}} & \multicolumn{2}{c|}{SketchyCOCO \cite{gao2020sketchyCOCO}} & \multicolumn{2}{c}{FSCOCO \cite{fscoco}} \\
         & & Acc.@1 & Acc.@10 & Acc.@1 & Acc.@10 \\\hline
        \multirow{2}{*}{\rotatebox{0}{\textbf{\makecell{S3}}}} & QST \cite{song2017textSketch} & 38.9 & 87.9 & 25.1 & 54.5 \\
         & SCM \cite{aytar2018crossmodal} & 38.5 & 87.3 & 24.3 & 54.1 \\\hline%\hdashline
        \multirow{1}{*}{\rotatebox{0}{\textbf{\makecell{B}}}} & CrossAtt \cite{transformer} & 39.1 & 88.2 & 25.3 & 54.8 \\\hline %\hdashline
        \rowcolor{Gray}
         & {\textbf{Proposed}} & \textbf{39.5} & \textbf{88.7} & \textbf{25.7} & \textbf{55.2} \\\bottomrule
    \end{tabular}
    \label{tab:stbir}
    \vspace{-0.2cm}
\end{table}
}

\begin{figure} [t]
    \centering
    \includegraphics[width=\linewidth]{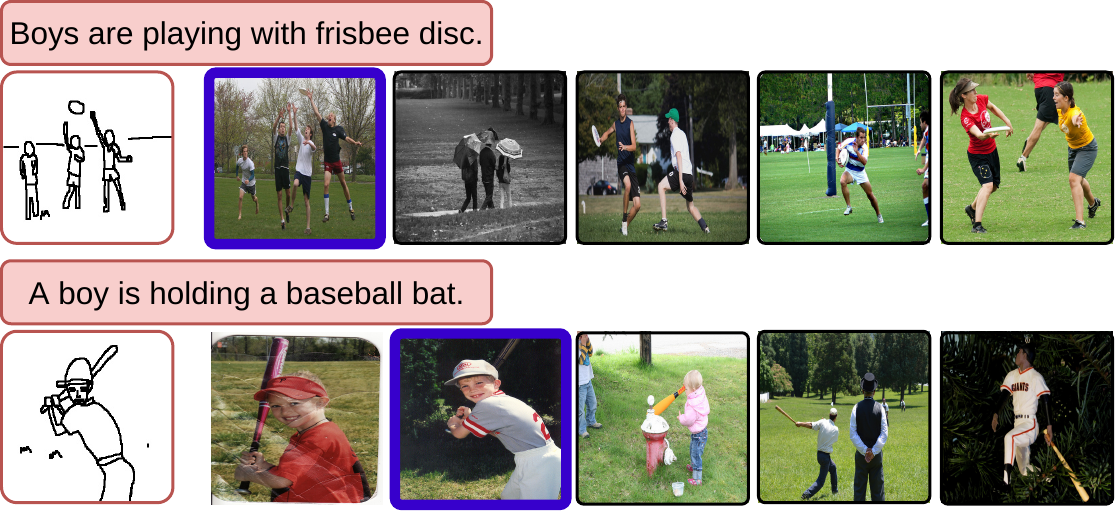}
    \vspace{-0.6cm}
    \caption{Qualitative results of combining sketch and text as \textcolor{sub-caption}{query} for image \textcolor{img-retrieve}{retrieval} on FSCOCO \cite{fscoco}. See supplementary for more.}
    \label{fig:sbir}
    \vspace{-0.4cm}
\end{figure}

\vspace{-0.1cm}
\subsection{Optionally using Sketch for Image Retrieval}
\vspace{-0.2cm}
\noindent Our method allows drawing only easy-to-sketch scenes instead of using both sketch and text forcibly. \Cref{tab:sbir} compares against methods that specialise on two-modalities (S2), three-modalities (S3), and our proposed baselines (B). We observe (i) training on three modalities (sketch, text, and photo) in S3 generally outperforms those trained using only sketch and photo (S2). This can be attributed to learning generalisable features in multi-modal setup \cite{aytar2018crossmodal}. (ii) \emph{QST} in S3 outperforms \emph{SCM} indicating quadruplet loss is a better training objective than naive element-wise addition when combining sketch, text, and photo. (iii) Performance difference between \emph{CrossAttn} and \emph{QST} is not as significant as in FG-STBIR (\Cref{tab:stbir}) as during inference, we only use sketch, omitting the cross-attention module. (iv) Our method outperforms S2, S3, and B even for two-modality setup thanks to disentanglement that eliminates confounding \cite{aytar2018crossmodal} modality-specific information.

{
\setlength{\tabcolsep}{5pt}
\begin{table}[t]
    \centering
    \footnotesize
    \caption{Quantitative results using only sketch for image retrieval (FG-SBIR) on two scene sketch datasets \cite{gao2020sketchyCOCO, fscoco}.}
    \vspace{-0.3cm}
    \begin{tabular}{clcc|cc}
        \toprule
        \multicolumn{2}{c}{\multirow{2}{*}{Method}} & \multicolumn{2}{c|}{SketchyCOCO \cite{gao2020sketchyCOCO}} & \multicolumn{2}{c}{FSCOCO \cite{fscoco}} \\
         & & Acc.@1 & Acc.@10 & Acc.@1 & Acc.@10 \\\hline
        \multirow{4}{*}{\rotatebox{0}{\textbf{\makecell{S2}}}} & Triplet-SN\cite{yu2016shoe} & 6.2 & 32.9 & 4.7 & 21.0 \\
         & HOLEF \cite{deep-spatial-semantic} & 6.2 & 40.7 & 4.9 & 21.7 \\
         & SketchyS \cite{zou2018sketchyscene} & 36.5 & 78.6 & 23.0 & 52.3 \\
         & SceneS \cite{liu2020scenesketcher} & 31.9 & 86.2 & -- & -- \\\hline %\hdashline
        \multirow{2}{*}{\rotatebox{0}{\textbf{\makecell{S3}}}} & QST \cite{song2017textSketch} & 37.4 & 87.1 & 23.6 & 52.9 \\
         & SCM \cite{aytar2018crossmodal} & 37.3 & 86.8 & 23.4 & 52.6 \\\hline %\hdashline
        \multirow{1}{*}{\rotatebox{0}{\textbf{\makecell{B}}}} & CrossAtt & 37.9 & 87.4 & 23.7 & 53.5 \\ \hline %\hdashline
        \rowcolor{Gray}
         & {\textbf{Proposed}} & \textbf{38.2} & \textbf{87.6} & \textbf{24.1} & \textbf{53.9} \\\bottomrule
    \end{tabular}
    \label{tab:sbir}
    \vspace{-0.1cm}
\end{table}
}

{
\begin{table}[!t]
    \centering
    \footnotesize
    \caption{Quantitative results of fine-grained text-based image retrieval (FG-TBIR) on two scene sketch datasets \cite{gao2020sketchyCOCO, fscoco}.}
    \vspace{-0.3cm}
    \begin{tabular}{clcc|cc}
        \toprule
        \multicolumn{2}{c}{\multirow{2}{*}{Method}} & \multicolumn{2}{c|}{SketchyCOCO \cite{gao2020sketchyCOCO}} & \multicolumn{2}{c}{FSCOCO \cite{fscoco}} \\
         & & Acc.@1 & Acc.@10 & Acc.@1 & Acc.@10 \\\hline
        \multirow{2}{*}{\rotatebox{0}{\textbf{\makecell{S2}}}} & CLIP \cite{CLIP} & 21.0 & 50.9 & 11.5 & 35.3 \\
         & CLIP-LN \cite{CLIP} & 22.1 & 52.3 & 14.8 & 36.6 \\\hline %\hdashline
        \multirow{2}{*}{\rotatebox{0}{\textbf{\makecell{S3}}}} & QST \cite{song2017textSketch} & 11.1 & 31.1 & 7.2 & 23.6 \\
         & SCM \cite{aytar2018crossmodal} & 10.7 & 31.0 & 6.9 & 23.1 \\ \hline %\hdashline
        \multirow{1}{*}{\rotatebox{0}{\textbf{\makecell{B}}}} & CrossAtt & 20.1 & 51.0 & 12.5 & 35.8 \\ \hline %\hdashline
        \rowcolor{Gray}
         & {\textbf{Proposed}} & \textbf{21.5} & \textbf{51.6} & \textbf{13.7} & \textbf{36.3} \\\bottomrule
    \end{tabular}
    \label{tab:tbir}
    \vspace{-0.1cm}
\end{table}
}

{
\setlength{\tabcolsep}{4.1pt}
\begin{table*}[!t]
    \centering
    \footnotesize
    \caption{Quantitative results of standard captioning metrics on MS-COCO \cite{mscoco-dataset} and FS-COCO \cite{fscoco} dataset.}
    \vspace{-0.3cm}
    \begin{tabular}{clcccccc|cccccc|cccccc}
        \toprule
        & & \multicolumn{6}{c}{Image Captioning} & \multicolumn{6}{c}{Sketch Captioning} & \multicolumn{6}{c}{Subjective Captioning} \\
        \multicolumn{2}{c}{Method} & B-1 & B-4 & M & R & C & S & B-1 & B-4 & M & R & C & S & B-1 & B-4 & M & R & C & S \\\hline
        \multirow{4}{*}{\rotatebox{90}{\textbf{S2}}} & SAT \cite{xu2015show} & 71.8 & 25.0 & 23.0 & -- & -- & -- & 46.2 & 13.7 & 17.1 & 44.9 & 69.4 & 14.5 & -- & -- & -- & -- & -- & -- \\
         & GMM-CVAE \cite{wang2017agcvae} & 72.9 & 30.7 & 24.2 & 52.5 & 98.6 & 17.7 & 49.6 & 15.5 & 18.3 & 48.7 & 77.6 & 15.5 & -- & -- & -- & -- & -- & --  \\
         & AG-CVAE \cite{wang2017agcvae} & 73.2 & 31.1 & 24.5 & 52.8 & 100.1 & 18.8 & 50.9 & 16.0 & 18.9 & 49.1 & 80.5 & 15.8 & -- & -- & -- & -- & -- & --  \\
         & LNFMM \cite{mahajan2020lnfmm} & 74.7 & 31.8 & 24.7 & 53.8 & 105.5 & 18.8 & 52.2 & 16.7 & 21.0 & 52.9 & 90.1 & 16.0 & -- & -- & -- & -- & -- & --  \\\hline %\hdashline
        \multirow{3}{*}{\rotatebox{90}{\textbf{B}}} & MulCap & 74.9 & 33.2 & 25.5 & 54.9 & 106.0 & 19.5 & 53.9 & 17.0 & 21.0 & 53.8 & 97.3 & 16.7 & 78.7 & 38.6 & 28.5 & 59.8 & 110.7 & 21.7 \\
         & CatCap & -- & -- & -- & -- & -- & -- & -- & -- & -- & -- & -- & -- & 77.6 & 38.0 & 28.3 & 57.7 & 108.0 & 21.2 \\
         & CrossCap & 75.5 & 34.3 & 26.1 & 55.4 & 106.7 & 20.1 & 54.3 & 17.9 & 21.4 & 54.3 & 100.3 & 17.5 & 79.2 & 39.3 & 28.4 & 59.5 & 117.3 & 22.1 \\\hline %\hdashline
        \rowcolor{Gray}
         & {\textbf{Proposed}} & \textbf{76.0} & \textbf{35.9} & \textbf{26.9} & \textbf{56.9} & \textbf{107.0} & \textbf{20.9} & \textbf{56.9} & \textbf{19.3} & \textbf{21.6} & \textbf{56.6} & \textbf{106.5} & \textbf{18.9} & \textbf{81.3} & \textbf{42.7} & \textbf{30.1} & \textbf{61.6} & \textbf{121.6} & \textbf{23.5} \\\bottomrule
    \end{tabular}
    \label{tab:image-sketch-captioning}
    \vspace{-0.3cm}
\end{table*}
}

\begin{figure} [t]
    \centering
    \includegraphics[width=\linewidth]{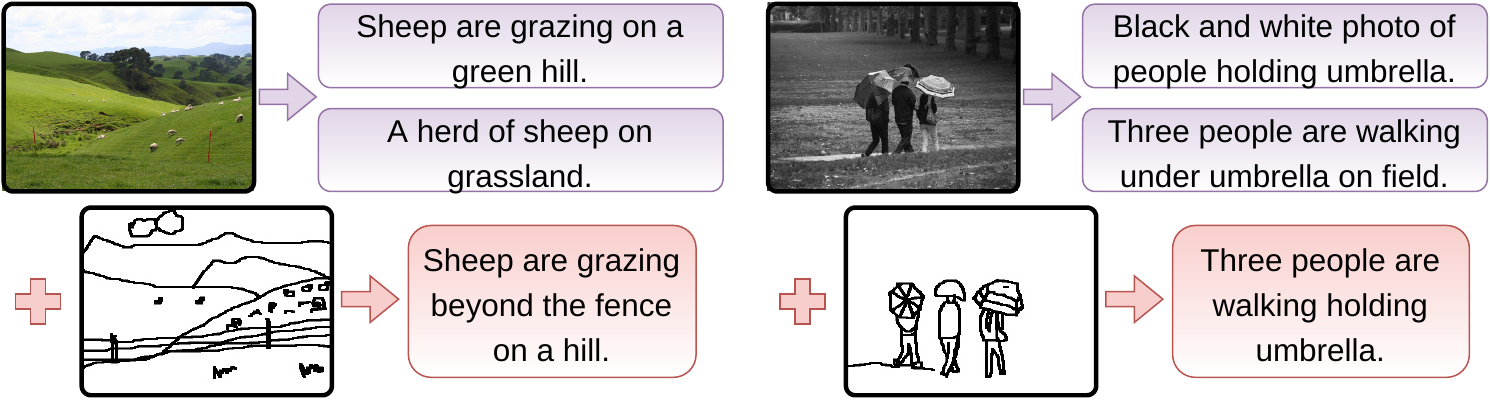}
    \vspace{-0.7cm}
    \caption{Qualitative results for \textcolor{img-caption}{image captioning} v/s \textcolor{sub-caption}{subjective captioning} on FS-COCO \cite{fscoco}. See supplementary for more.}
    \label{fig:captioning}
    \vspace{-0.4cm}
\end{figure}

{
\setlength{\tabcolsep}{8.3pt}
\begin{table}[!htbp]
    \centering
    \footnotesize
    \caption{Ablation study on FG-STBIR and Subjective Captioning using FSCOCO \cite{fscoco}. \texttt{CA} denotes cross-attention in \cref{sec: set-attention}.}
    \vspace{-0.3cm}
    \begin{tabular}{ccccccc}
        \toprule
        $\tau_{\mathbf{k}}$ & \texttt{CA} & $\mathcal{L}_{cl}$ & Acc.@1 & Acc.@10 & B-1 & C \\\hline
        \xmark & \xmark & \xmark & 24.5 & 53.7 & 73.3 & 100.1 \\
        \cmark & \xmark & \xmark & 24.9 & 54.0 & 77.9 & 108.5 \\
        \cmark & \cmark & \xmark & 25.5 & 54.9 & 80.6 & 119.3 \\
        \rowcolor{Gray}
        \cmark & \cmark & \cmark & \textbf{25.7} & \textbf{55.2} & \textbf{81.3} & \textbf{121.6} \\\bottomrule
    \end{tabular}
    \label{tab:ablation}
    \vspace{-0.5cm}
\end{table}
}

\vspace{-0.1cm}
\subsection{Optionally using Text for Image Retrieval}
\vspace{-0.2cm}
\noindent While some information is best expressed by drawing, others, like colour, is best described via text. From \Cref{tab:tbir}, we  observe (i) Given the same train/test split, sketches outperform text as a query modality for fine-grained image retrieval. (ii) \emph{CLIP} and \emph{CLIP-LN} outperforms all competitors due to superior pre-trained weights using $400$ million text-image pairs. (iii) The proposed method outperforms most methods due to disentanglement that drives out modality-specific components. Although CLIP \cite{CLIP} outperforms the proposed method, we deliberately use a simple and easy-to-reproduce GRU/VGG-16 architectures for text/photo encoders, and train on a much smaller data \cite{fscoco, gao2020sketchyCOCO} than CLIP.

\vspace{-0.1cm}
\subsection{Image or Sketch Captioning}
\vspace{-0.2cm}
\noindent In addition to disentanglement for cross-modal retrieval tasks (e.g., FG-SBIR, FG-TBIR), our conditional invertible neural network $\tau_{\mathbf{t}}$ can also generate text-specific information (\cref{fig:cINN-schematic}) to support generative tasks like image/sketch captioning. We generate $100$ candidate captions using (i) beam search for \emph{SAT}, \emph{MulCap}, \emph{CrossCap}, \emph{CatCap}, and (ii) sampling from prior distribution for \emph{GMM-CVAE}, \emph{AG-CVAE}, \emph{LNFMM}, and our proposed method. From \Cref{tab:image-sketch-captioning}, we observe (i) our baselines adopting recent techniques like vision-transformer \cite{VIT} outperforms (S2) -- recent but complex approaches like \emph{LNFMM}, \emph{AG-CVAE}, and the older yet seminal work like \emph{SAT}. (ii) Performance gap between \emph{MulCap} and \emph{CrossCap} is insignificant for two-modality setups (photo to text, or sketch to text) since they primarily differentiate by their multi-modal (photo and sketch) fusion strategy. (iii) In spite of using a photo/sketch encoder and text decoder similar to our simple competitor \emph{SAT}, our proposed method performs competitively with complex methods like \emph{LNFMM}, \emph{AG-CVAE}, and latest approaches using vision-transformers \cite{VIT}, like \emph{CrossCap}. This shows the significant contribution of (i) disentangling modality-specific and modality-agnostic components from photo/sketch, and (ii) modelling text-specific prior for generative tasks.

\vspace{-0.1cm}
\subsection{Sketch Based Subjective Captioning}
\vspace{-0.2cm}
\noindent As defined in \cref{sec: pilot-captioning}, unlike traditional captioning frameworks that factually describe an image or sketch in a neutral tone, subjective captioning focus on drawing out a user's intentions, salient objects, and artistic interpretations \cite{caricatureshop}. Being the first method to use scene-level sketch as a guiding signal for captioning, we follow controllable captioning literature \cite{stefanini2021captioningsurvey} to adopt three baselines (\textbf{B}) that inject the sketch conditioning signal into the captioning pipeline. From \Cref{tab:image-sketch-captioning}, we observe (i) \emph{MulCap} outperforms \emph{CatCap}, thereby supporting previous observations \cite{changpinyo2019captioning} of element-wise multiplication being more effective than concatenation. (ii) \emph{CrossAtt} outperforms all baselines (B) and two-modality SOTAs (S2) by using a cross-attention mechanism to fuse sketch and photo by modelling sketch-photo interactions to resolve overlapping or conflicting information. Our proposed method is similar to \emph{CrossAtt} using cross-attention (\cref{sec: set-attention}) but also enriches the modality-agnostic sketch and photo features by removing the confounding modality-specific information to offer the best performance.

\vspace{-0.1cm}
\subsection{Ablation}
\vspace{-0.2cm}
\noindent In \Cref{tab:ablation}, we evaluate the contribution of each key design choice on FG-STBIR and Subjective captioning using FS-COCO \cite{fscoco}. (i) Replacing cross-attention in \cref{sec: set-attention} with quadruplet loss \cite{song2017textSketch} leads to a performance drop by $0.6/0.9/2.7/10.8$ in Acc.@1/Acc.@10/B-1/C metrics respectively to show the importance of modelling the interaction between sketch and text. (ii) Replacing contrastive loss-based query-photo score in \cref{eq: contrastive} with a simple triplet loss leads to a performance drop by $0.2/0.3/0.7/2.3$ due to the inability of unimodal $L2$-based triplet loss to model highly complex scene information \cite{cpc2018}. (iii) Finally, removing the conditional invertible neural networks ($\tau_{\mathbf{k}}$) drops retrieval and captioning by $0.4/0.3/4.6/8.4$ due to percolation of the confounding modality-specific information in cross-modal tasks \cite{aytar2018crossmodal} and the inability to generate text-specific information from photo and sketch respectively.
% \vspace{-0.3cm}

\vspace{-0.1cm}
\section{Conclusion}
\vspace{-0.2cm}
\noindent We have studied for the first time the trilogy relationship among scene-level sketch, text, and photo by introducing scene-sketch in the context of scene understanding. We proposed a unified framework to jointly model sketch, text, and photo that seamlessly support several downstream tasks like fine-grained sketch-based image retrieval, fine-grained sketch and text based image retrieval, sketch captioning, and subjective captioning, among others. Future research can explore challenging downstream tasks such as scene-level sketch-based image generation, sketch and text based image generation, and text-based sketch generation tasks.

\appendix

\definecolor{commentcolor}{RGB}{110,154,155}   % define comment color
\newcommand{\PyComment}[1]{\ttfamily\textcolor{commentcolor}{\# #1}}  % add a "#" before the input text "#1"
\newcommand{\PyCode}[1]{\ttfamily\textcolor{black}{#1}} % \ttfamily is the code font

\section{Details for Subjective Captioning}
We provide additional details of our pilot study in Sec.~\textcolor{red}{3.2} that compare the performance of subjective captioning when using part-of-speech (POS) \cite{deshpande2019pos}, mouse trace \cite{meng2021mouse} or sketch as a guiding signal into the image captioning pipeline. Instead of choosing a common baseline to compare subjective captioning when using POS, mouse trace, and sketches, we measure the relative performance over the standard baselines used in recent literature to study the contribution of every guiding signal.
(i) For POS \cite{deshpande2019pos}, we measure the relative performance using Wang \etal \cite{wang2017agcvae} as baseline. Without using POS, i.e., (w/o)-POS gives a B-4/C score of $31.1/100$ as compared to with POS, i.e., (w)-POS that gives $31.6/104/5$. 
(ii) For mouse trace \cite{deshpande2019pos}, we use \cite{pont2020mouse} to get (w/o)-Trace B-4/C score of $8.1/29.3$ as compared to (w)-Trace score of $24.6/106.5$. This leads to a large relative improvement of $16.5/77.2$ to show the significant contribution of using mouse trace as guiding signal.
(iii) For sketch, we follow \cite{fscoco} to use \cite{mahajan2020lnfmm} as baseline to get (w/o)-Sketch B-4/C score of $31.8/42.7$. We use cross-attention mechanism in \cite{meng2021mouse} to inject sketch as a guiding signal into our baseline \cite{mahajan2020context} to give a (w)-Sketch score of $42.7/121.6$. This gives a relative improvement of $10.9/16.1$, which shows that sketch as a guiding signal is better than POS and competitive as mouse trace. Hence, we advocate for sketch as a guiding signal to depict saliency since unlike POS \cite{deshpande2019pos} or mouse trace \cite{meng2021mouse}, sketches are more expressive that can capture artistic interpretation like caricature \cite{caricatureshop}.

\section{Modelling more than three modalities}
Sec.~\textcolor{red}{4.4} optionally models the modality-agnostic components of sketch or text using the function $\mathcal{G}(\cdot)$ that consists of a multihead cross-attention module $\texttt{MH}(\cdot)$ followed by an attention-based pooling $\texttt{PMA}(\cdot)$. For $M=3$, $\mathcal{L}_{cls}^{tot}$ is defined as,
\begin{equation}
    \begin{split}
        \mathcal{L}_{cl}^{tot} =& \mathcal{L}_{cl}(\mathcal{G}(f^{ag}_{\mathbf{s}}, f^{ag}_{\mathbf{t}}), f^{ag}_{\mathbf{p}})  \\
        & \hspace{-2em} + \mathcal{L}_{cl}(\mathcal{G}(f^{ag}_{\mathbf{s}}, f^{ag}_{\mathbf{p}}), f^{ag}_{\mathbf{t}}) + \mathcal{L}_{cl}(\mathcal{G}(f^{ag}_{\mathbf{p}}, f^{ag}_{\mathbf{t}}), f^{ag}_{\mathbf{s}})
    \end{split}
\end{equation}
In this section, we show how $\mathcal{G}(\cdot)$ can be extended to more than three modalities $M>3$. Given a set of modality-agnostic components as $\Psi = \{f^{ag}_{\mathbf{1}}, f^{ag}_{\mathbf{2}}, \dots, f^{ag}_{\mathbf{M}} \}$, we can solve for $\mathcal{L}^{tot}_{cl}$ as,
\begin{equation}\label{eq: generic-contrastive-supp}
    \mathcal{L}^{tot}_{cl} = \sum_{j=1}^{\mathbf{M}} \mathcal{L}_{cl}( \mathcal{G}( \Psi - \{f^{ag}_{{j}}\} ), f^{ag}_{{j}} )
\end{equation}
We further elaborate \cref{eq: generic-contrastive-supp} using \cref{alg: generic-contrastive-supp}.

\begin{algorithm}
    \caption{Compute generalised $\mathcal{L}^{tot}_{cl}$ for $M>3$}\label{alg: set-attention}
    \begin{algorithmic}
        \Require $\mathcal{P} \in \mathbb{R}^{1 \times 480}$ \Comment Learned weights.
        \State $\Psi = \{f^{ag}_{\mathbf{1}}, f^{ag}_{\mathbf{2}} \dots, f^{ag}_{\mathbf{M}} \}, \ \in \mathbb{R}^{M \times 480}$
        \State $\mathcal{L}^{tot}_{cl} \gets 0$
        \For{$j \gets 1 \text{ to } M$}
            \State $\mathcal{S}_{M} \gets \Psi - \{f^{ag}_{\mathbf{i}} \}$ \Comment ${(M-1) \times 480}$
            \State $H_{M} \gets \texttt{MH}(\mathcal{S}_{M})$ \Comment ${(M-1) \times 480}$
            \State $f_{M} = \texttt{PMA}(H_{M}) = \sigma(\mathcal{P} H_{M}^{T}) H_{M}$ \Comment $({1 \times 480} )$
            \State $\mathcal{L}^{tot}_{cl} \gets \mathcal{L}^{tot}_{cl} + \mathcal{L}_{cl} (f^{ag}_{{j}}, f_{M})$
        \EndFor
        \State \Return $\mathcal{L}^{tot}_{cl}$
    \end{algorithmic}
    \label{alg: generic-contrastive-supp}
\end{algorithm}

\section{Derivation of Disentanglement Loss in Eq.~\textcolor{red}{3}}

For optionality across tasks, we disentangle the information from sketch, text, and photo, given by $\mathbf{k} \in \{ \mathbf{s}, \mathbf{t}, \mathbf{p} \}$ into a discriminative part $f^{ag}_{\mathbf{k}}$ shared across modalities, and a generative part specific to one modality $f^{sp}_{\mathbf{k}}$. This information split of $f_{\mathbf{k}} = [ f^{ag}_{\mathbf{k}}, f^{sp}_{\mathbf{k}}]$ is achieved in Sec.~\textcolor{red}{4.3} by minimising the mutual information between the modality-agnostic and modality-specific components defined as,
% \vspace{-0.3cm}
\begin{equation}\label{eq: mutual-disentangle-supp}
    \begin{split}
        \hspace{-0.5em} \mathcal{I}(f^{ag}_{\mathbf{k}}, f^{sp}_{\mathbf{k}}) = & \int_{f^{ag}_{\mathbf{k}}, f^{sp}_{\mathbf{k}}} \mathbbm{p} (f^{ag}_{\mathbf{k}}, f^{sp}_{\mathbf{k}}) \log \frac{ \mathbbm{p} (f^{ag}_{\mathbf{k}}, f^{sp}_{\mathbf{k}}) }{ \mathbbm{p}(f^{ag}_{\mathbf{k}}) \mathbbm{p}(f^{sp}_{\mathbf{k}}) } \\
         & = \int_{f^{ag}_{\mathbf{k}}, f^{sp}_{\mathbf{k}}} \mathbbm{p} (f^{ag}_{\mathbf{k}}, f^{sp}_{\mathbf{k}}) \log \frac{ \mathbbm{p} (f^{sp}_{\mathbf{k}} | f^{ag}_{\mathbf{k}}) }{ \mathbbm{p}(f^{sp}_{\mathbf{k}}) }
    \end{split}
\end{equation}
Given a variational distribution $\mathbbm{q}(f^{sp}_{\mathbf{k}})$, due to positivity of KL divergence we have,
% \vspace{-0.1cm}
\begin{equation}
    \int \mathbbm{p}(f^{sp}_{\mathbf{k}}) \log \mathbbm{p} (f^{sp}_{\mathbf{k}}) \geq \int \mathbbm{p}(f^{sp}_{\mathbf{k}}) \log \mathbbm{q} (f^{sp}_{\mathbf{k}})
\end{equation}
Hence, approximating the modality-specific prior $\mathbbm{p}(f^{sp}_{\mathbf{k}})$ with variational distribution $\mathbbm{q}(f^{sp}_{\mathbf{k}})$ in \cref{eq: mutual-disentangle-supp} we get,
% \vspace{-0.3cm}
\begin{equation}
    \mathcal{I}(f^{ag}_{\mathbf{k}}, f^{sp}_{\mathbf{k}}) \leq \int_{f^{ag}_{\mathbf{k}}, f^{sp}_{\mathbf{k}}} \mathbbm{p} (f^{ag}_{\mathbf{k}}, f^{sp}_{\mathbf{k}}) \log \frac{ \mathbbm{p} (f^{sp}_{\mathbf{k}} | f^{ag}_{\mathbf{k}}) }{ \mathbbm{q}(f^{sp}_{\mathbf{k}}) }
\end{equation}
Assuming a uniform prior distribution $\mathbbm{p}(\eta)$, and its definition in Eq.~\textcolor{red}{2} via conditional invertible neural network $\tau_{\mathbf{k}}$, we have,
\begin{equation}
\begin{split}
    \mathcal{L}_{\tau_{\mathbf{k}}} = & - \mathbb{E}_{f^{sp}_{\mathbf{k}}, f^{ag}_{\mathbf{k}}} \{ \log \mathbbm{q}(\tau^{-1}_{\mathbf{k}}(f^{sp}_{\mathbf{k}} \ | \ f^{ag}_{\mathbf{k}})) \\
     & + \log|\mathrm{det} J_{\tau^{-1}_{\mathbf{k}}}(f^{sp}_{\mathbf{k}} \ | \ f^{ag}_{\mathbf{k}})| \} - H(f^{sp}_{\mathbf{k}} | f^{ag}_{\mathbf{k}})
\end{split}
\end{equation}
where, $H(f^{sp}_{\mathbf{k}} | f^{ag}_{\mathbf{k}})$ is the constant data entropy which is ignored in the final optimisation in Eq.~\textcolor{red}{3}.

\section{Comparison with a parallel work \cite{sangkloy2022textSketch}}
A parallel work surfaced while writing this paper by Sangkloy \etal \cite{sangkloy2022textSketch} can optionally perform text-based image retrieval (TBIR), sketch-based image retrieval (SBIR), or both sketch+text based image retrieval (STBIR). However, the motivation of \cite{sangkloy2022textSketch} is crucially different from ours. While we focus on improving the latent space via disentanglement into a modality-specific and modality-agnostic component to support optionality across tasks (retrieval and captioning) and modalities (using only sketch, only text, or both as query), Sangkloy \etal \cite{sangkloy2022textSketch} focused on improving the encoders for sketch, text, and photo by adapting the recently popular pre-trained CLIP \cite{CLIP}. To model only sketch, only text, or both sketch+text for image retrieval, \cite{sangkloy2022textSketch} used a rather simple late-fusion technique performing element-wise addition of sketch and text features. While the training code of the proposed model in \cite{sangkloy2022textSketch} is not been released yet, our re-implementation of \cite{sangkloy2022textSketch} using simple element-wise addition of sketch and text features with CLIP encoders lead to STBIR performance of $23.9/53.5$ in Acc.@1/Acc.@10 which is significantly lower than our proposed method by $15.6/35.2$ on FS-COCO \cite{fscoco}. Although CLIP \cite{CLIP} is highly generalisable to open-set setups, it is difficult to adapt to small downstream datasets like FS-COCO \cite{fscoco} and simultaneously outperform task-specific encoders like VGG-16 \cite{vgg-16} used in the proposed method. A similar trend was also observed in Chowdhury \etal \cite{fscoco}.

\section{Clarification on Contributions}
Our goal is not to design a model that is state-of-the-art for ALL retrieval (e.g., FG-STBIR, FG-SBIR, FG-TBIR) and generative (e.g., image, sketch, and subjective captioning) tasks. Instead, we (i) design a generalisable model that is competitive with a myriad of baselines (large models like CLIP-LN or small ones like VGG) across multiple tasks; (ii) we show how the benefits of sketch modality (acknowledged by several prior works \cite{fscoco, tripathi2020object}) can be optionally combined with multiple modalities like text and photo.

\section{Comparison with Matrix Factorization}
While our baseline MulCap performs feature multiplication similar to matrix factorization \cite{veit2018separating, llerena2019cross-modality}, we additionally adopt \cite{veit2018separating} to get subjective captioning (BELU-1, CIDEr) score of ($79.2 \pm 0.6$, $113.5 \pm 1.1$).

\section{Evaluation with different training seeds}
Training on $5$ different seeds, we report accuracy on FG-STBIR task. For FS-COCO \cite{fscoco} we get Acc.@1 and Acc.@10 of $25.6 \pm 0.5$ and $55.3 \pm 0.3$ respectively. Further experimenting on shoe dataset \cite{yu2016shoe}, we get FG-STBIR Acc.@1 and Acc.@10 scores of $53.2 \pm 0.5$ and $88.1 \pm 0.2$.

\section{Additional Details on Pilot Study}
Our pilot study aims to: (i) compare sketch vs. text as a query for fine-grained image retrieval. For this, we use standard baselines Triplet-SN (for SBIR) and CLIP-LN (for TBIR) on $3000$ sketch/photo, and text/photo pairs in FS-COCO \cite{fscoco}. We observe that for some instances sketch is a better query for image retrieval as it can depict complex shapes, multiple objects, and spatial alignment. However, not all objects are easy to draw (e.g., differentiate a `donkey' vs. a `horse') but could be easily described via text. (ii) For subjective captioning, we compare the relative improvements in standard captioning metrics (like M, R, C, S) when using users' sketch (to generate subjective captions) vs. without using sketches (to generate subjective captions).

\section{Comparison with Aytar \etal \cite{aytar2017see}}
Ayatar \etal \cite{aytar2017see} learns a joint embedding space across image, sound, and text. This is similar to our method, which also aims to learn a joint embedding space across image, sketch, and text. However, there are some key differences: (i) \cite{aytar2017see} lacks the ability to combine multiple modalities like sound+text for image retrieval. The ability to optionally combine multiple modalities for image retrieval is crucial to our motivation, e.g., fine-grained sketch-based image retrieval (FG-SBIR), fine-grained text-based image retrieval (FG-TBIR), and fine-grained sketch+text based image retrieval (FG-STBIR). (ii) The embedding space of \cite{aytar2017see} only supports discriminative tasks. This fails to support the generative objectives of our method, like image captioning, sketch captioning, and subjective captioning. Nevertheless, we compare Acc.@1 with \cite{aytar2017see} on FS-COCO \cite{fscoco} for FG-SBIR and FG-TBIR to get $23.5\%$ and $7.1\%$ respectively.

\section{Differences from prior works}
Prior works like (i) Aytar \etal \cite{aytar2018crossmodal} study only cross-modal transfer between a pair of modalities (sketch/photo, or text/photo), not a combination of multiple modalities (sketch+text, or sketch+photo) nor feature disentanglement (modality-agnostic and modality-specific) which is crucial for tasks like FG-STBIR and subjective captioning. (ii) Song \etal \cite{song2017textSketch} combines sketch+text for image retrieval via a weighted sum of sketch-photo and text-photo distances computed independently. This simple setup is (a) limited to retrieval (i.e., no captioning), and (b) lacks feature disentanglement to filter our irrelevant modality-specific information (drawing style) when combining multiple modalities (sketch+text). We bring new insights into scene understanding by showing the need for feature disentanglement to (i) optionally combine multiple modalities, and (ii) support both discriminative and generative tasks.

\section{Complex Faliure Cases}

We show qualitative results below where sketch + text performs poorly. We observe this happens when both the input sketch or text is ambigious (i.e., badly drawn sketch or unprecise short textual phrases).
\begin{figure}[!h]
    \centering
    \includegraphics[width=\linewidth]{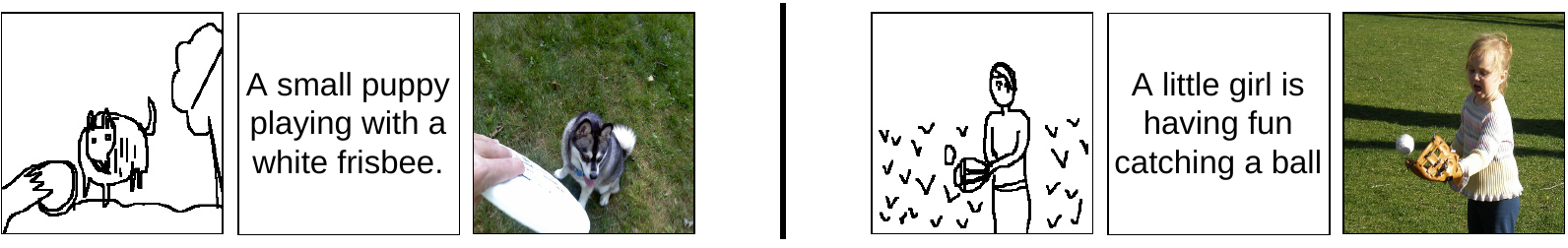}
    % \caption{Qualitative results showing complex failure cases where sketch+text performs poorly.}
\end{figure}

\section{Sketch+Text as Query for Image Retrieval}

\begin{figure}[!h]
    \centering
    \includegraphics[width=\linewidth]{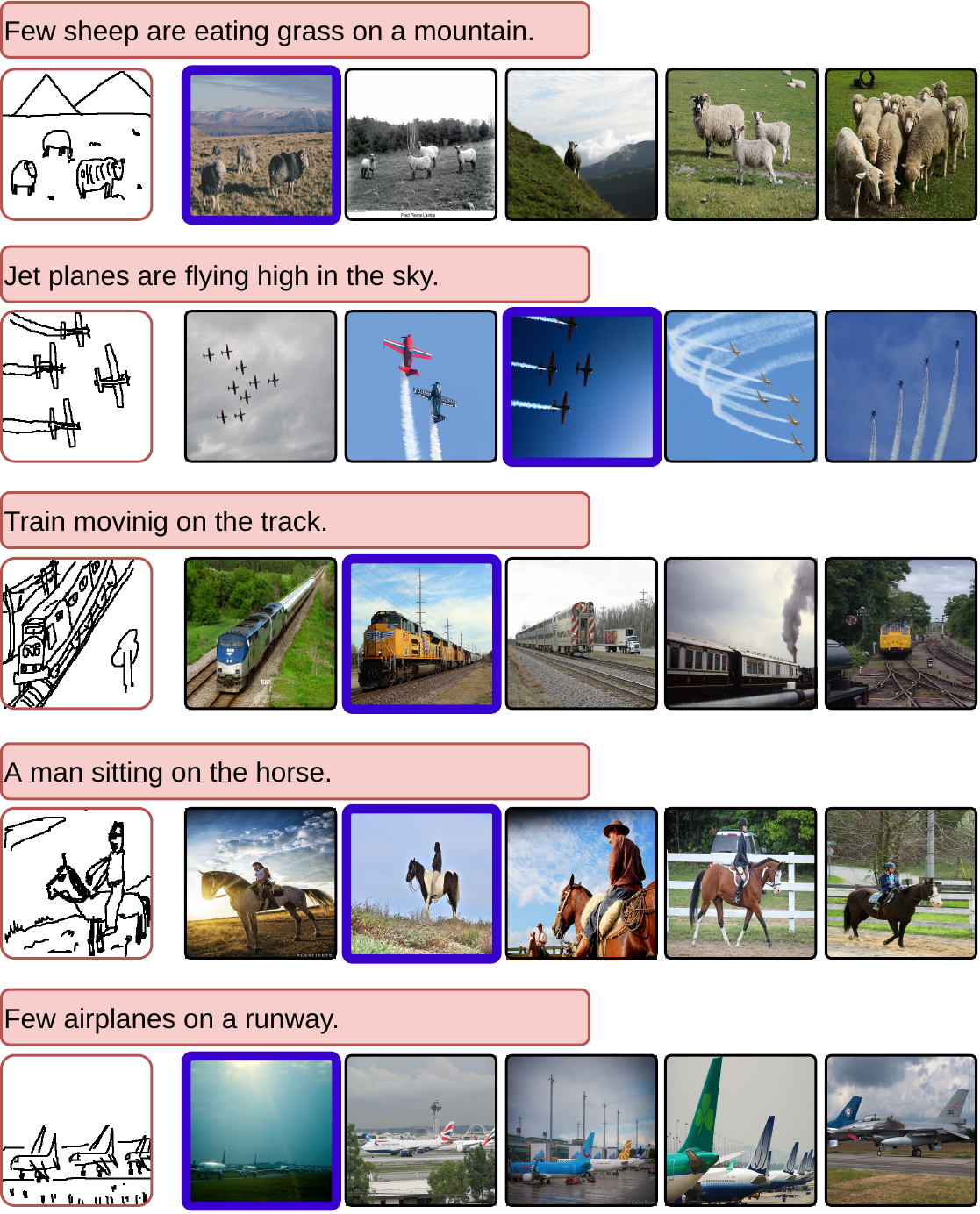}
\end{figure}

\newpage

\section{Image vs. Subjective Captioning}

\begin{figure}[!h]
    \centering
    \includegraphics[width=\linewidth]{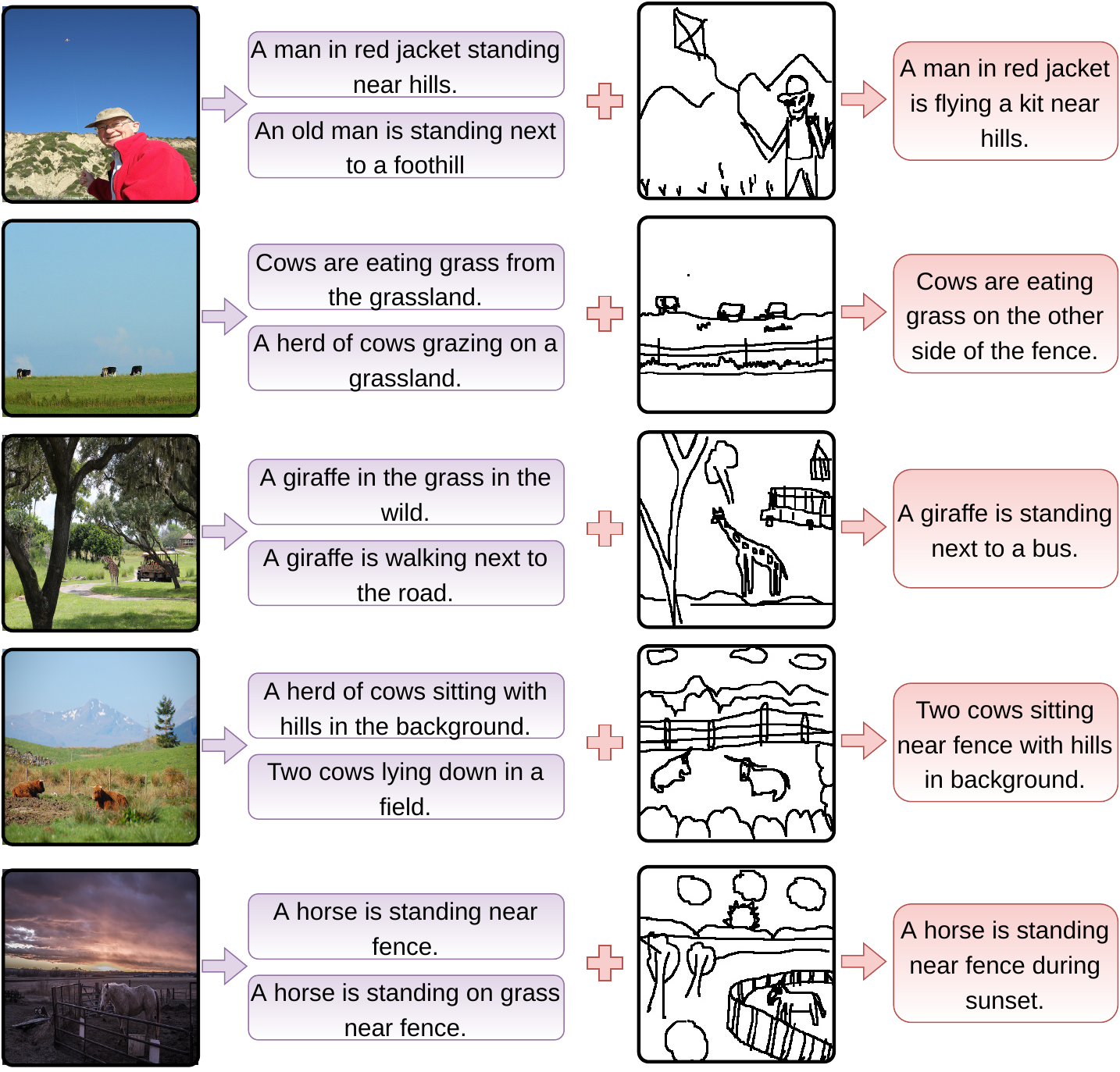}
\end{figure}

\newpage

%%%%%%%%% REFERENCES
{\small
\bibliographystyle{ieee_fullname}
\bibliography{camera_ready}

\begin{thebibliography}{100}\itemsep=-1pt

\bibitem{agarwal2019nocaps}
Harsh Agarwal, Karan Desai, Yufei Wang, Xinlei Chen, Rishabh Jain, Mark
  Johnson, Dhruv Batra, Devi Parikh, Stefan Lee, and Peter Anderson.
\newblock nocaps: novel object captioning at scale.
\newblock In {\em ICCV}, 2019.

\bibitem{anderson2016spice}
Peter Anderson, Basura Fernando, Mark Johnson, and Stephen Gould.
\newblock Spice: semantic propositional image caption evaluation.
\newblock In {\em ECCV}, 2016.

\bibitem{aytar2018crossmodal}
Yusuf Aytar, Lluis Castrejon, Carl Vondrick, Hamed Pirsiavash, and Antonio
  Torralba.
\newblock Cross-modal scene networks.
\newblock {\em IEEE TPAMI}, 2018.

\bibitem{aytar2017see}
Yusuf Aytar, Carl Vondrick, and Antonio Torralba.
\newblock {See, Hear, and Read: Deep Aligned Representations}.
\newblock {\em arXiv preprint arXiv:1706.00932}, 2017.

\bibitem{layer-norm}
Jimmy~Lei Ba, Jamie~Ryan Kiros, and Geoffrey~E. Hinton.
\newblock Layer normalization.
\newblock {\em arXiv preprint arXiv:1607.06450}, 2016.

\bibitem{artcaptioning}
Zechen Bai, Yuta Nakashima, and Noa Gracia.
\newblock Explain me the painting: Multi-topic knowledgeable art description
  generation.
\newblock In {\em ICCV}, 2021.

\bibitem{bhunia2021semi}
Ayan~Kumar Bhunia, Pinaki~Nath Chowdhury, Aneeshan Sain, Yongxin Yang, Tao
  Xiang, and Yi-Zhe Song.
\newblock More photos are all you need: Semi-supervised learning for
  fine-grained sketch based image retrieval.
\newblock In {\em CVPR}, 2021.

\bibitem{vector-raster}
Ayan~Kumar Bhunia, Pinaki~Nath Chowdhury, Yongxin Yang, Timothy~M. Hospedales,
  Tao Xiang, and Yi-Zhe Song.
\newblock Vectorization and rasterization: Self-supervised learning for sketch
  and handwriting.
\newblock In {\em CVPR}, 2021.

\bibitem{pixelor}
Ayan~Kumar Bhunia, Ayan Das, Umar~Riaz Muhammad, Yongxin Yang, Timothy~M.
  Hospedales, Tao Xiang, Yulia Gryaditskaya, and Yi-Zhe Song.
\newblock Pixelor: A competitive sketching ai agent. so you think you can beat
  me?
\newblock {\em ACM TOG}, 2020.

\bibitem{bhunia2022worrying}
Ayan~Kumar Bhunia, Subhadeep Koley, Abdullah Faiz Ur~Rahman Khilji, Aneeshan
  Sain, Pinaki~Nath Chowdhury, Tao Xiang, and Yi-Zhe Song.
\newblock Sketching without worrying: Noise-tolerant sketch-based image
  retrieval.
\newblock In {\em CVPR}, 2022.

\bibitem{bhunia2023sketch2saliency}
Ayan~Kumar Bhunia, Subhadeep Koley, Amandeep Kumar, Aneeshan Sain, Pinaki~Nath
  Chowdhury, Tao Xiang, and Yi-Zhe Song.
\newblock {Sketch2Saliency: Learning to Detect Salient Objects from Human
  Drawings}.
\newblock In {\em CVPR}, 2023.

\bibitem{bhunia2022adaptive}
Ayan~Kumar Bhunia, Aneeshan Sain, Parth Shah, Animesh Gupta, Pinaki~Nath
  Chowdhury, Tao Xiang, and Yi-Zhe Song.
\newblock Adaptive fine-grained sketch-based image retrieval.
\newblock In {\em ECCV}, 2022.

\bibitem{bhunia2020sketch}
Ayan~Kumar Bhunia, Yongxin Yang, Timothy~M. Hospedales, Tao Xiang, and Yi-Zhe
  Song.
\newblock Sketch less for more: On-the-fly fine-grained sketch based image
  retrieval.
\newblock In {\em CVPR}, 2020.

\bibitem{castrejon2016cmplaces}
Ll. Castrejón, Y. Aytar, C. Vondrick, H. Pirsiavash, and A. Torralba.
\newblock Learning aligned cross-modal representations from weakly aligned
  data.
\newblock In {\em CVPR}, 2016.

\bibitem{changpinyo2019captioning}
Soravit Changpinyo, Bo Pang, Piyush Sharma, and Radu Soricut.
\newblock Decoupled box proposal and featurization with ultrafine-grained
  semantic labels improve image captioning and visual question answering.
\newblock In {\em EMNLP}, 2019.

\bibitem{chen2021captioningverbs}
Long Chen, Zhihong Jiang, Jun Xiao, and Wei Liu.
\newblock Human-like controllable image captioning with verb-specific semantic
  roles.
\newblock In {\em CVPR}, 2021.

\bibitem{infoGAN}
Xi Chen, Yan Duan, Rein Houthooft, John Schulman, Ilya Sutskever, and Pieter
  Abbeel.
\newblock Infogan: interpretable representation learning by information
  maximizing generative adversarial nets.
\newblock In {\em NeurIPS}, 2016.

\bibitem{partially-does-it}
Pinaki~Nath Chowdhury, Ayan~Kumar Bhunia, Viswanatha~Reddy Gajjala, Aneeshan
  Sain, Tao Xiang, and Yi-Zhe Song.
\newblock {Partially Does It:} towards scene-level {FG-SBIR} with partial
  input.
\newblock In {\em CVPR}, 2022.

\bibitem{chowdhury2023detect}
Pinaki~Nath Chowdhury, Ayan~Kumar Bhunia, Aneeshan Sain, Subhadeep Koley, Tao
  Xiang, and Yi-Zhe Song.
\newblock {What Can Human Sketches Do for Object Detection?}
\newblock In {\em CVPR}, 2023.

\bibitem{fscoco}
Pinaki~Nath Chowdhury, Aneeshan Sain, Yulia Gryaditskaya, Ayan~Kumar Bhunia,
  Tao Xiang, and Yi-Zhe Song.
\newblock Fs-coco: Towards understanding of freehand sketches of common objects
  in context.
\newblock In {\em ECCV}, 2022.

\bibitem{chowdhury20223Dsynthesis}
Pinaki~Nath Chowdhury, Tuanfeng Wang, Duygu Ceylan, Yi-Zhe Song, and Yulia
  Gryaditskaya.
\newblock Garment ideation: Iterative view-aware sketch-based garment modeling.
\newblock In {\em 3DV}, 2022.

\bibitem{chun2021pcme}
Sanghyuk Chun, Joon Oh, Seong, Sampaio~de Rezende, Yannis Kalantidis, and Diane
  Larlus.
\newblock Probabilistic embeddings for cross-modal retrieval.
\newblock In {\em CVPR}, 2021.

\bibitem{liveSketch}
John Collomosse, Tu Bui, and Jin Hailin.
\newblock Livesketch: Query perturbations for guided sketch-based visual
  search.
\newblock In {\em CVPR}, 2019.

\bibitem{cornia2020m2net}
Marcella Cornia, Matteo Stefanini, Lorenzo Baraldi, and Rita Cucchiara.
\newblock Meshed-memory transformer for image captioning.
\newblock In {\em CVPR}, 2020.

\bibitem{imagenet}
Jia Deng, Wei Dong, Richard Socher, Li-Jia Li, Kai Li, and Li Fei-Fei.
\newblock Imagenet: A large-scale hierarchical image database.
\newblock In {\em CVPR}, 2009.

\bibitem{denkowski2014meteor}
Michael~J. Denkowski and Alon Lavie.
\newblock Meteor universal: Language specific translation evaluation for any
  target language.
\newblock In {\em WMT@ACL}, 2014.

\bibitem{deshpande2019pos}
Aditya Deshpande, Jyoti Aneja, Liwei Wang, Alexander Schwing, and D.~A.
  Forsyth.
\newblock Fast, diverse and accurate image captioning guided by part-of-speech.
\newblock In {\em CVPR}, 2019.

\bibitem{doodle-to-search}
Sounak Dey, Pau Riba, Anjan Dutta, Josep Llados, and Yi-Zhe Song.
\newblock Doodle to search: Practical zero-shot sketch-based image retrieval.
\newblock In {\em CVPR}, 2019.

\bibitem{nice}
Laurent Dinh, David Krueger, and Yoshua Bengio.
\newblock Nice: Non-linear independent components estimation.
\newblock In {\em ICLR}, 2015.

\bibitem{realNVP}
Laurent Dinh, Jascha Sohl-Dickstein, and Samy Bengio.
\newblock Density estiamtion using real nvp.
\newblock In {\em ICLR}, 2017.

\bibitem{VIT}
Alexey Dosovitskiy, Lucas. Beyer, Alexander Kolesnikov, Dirk Weissenborn,
  Xiaohua Zhai, Thomas Unterthiner, Mostafa Dehghani, Matthias Minderer, Georg
  Heigold, Sylvain Gelly, Jakob Uszkoreit, and Neil Houlsby.
\newblock An image is wort 16x16 words: Transformers for image recognition at
  scale.
\newblock In {\em ICLR}, 2021.

\bibitem{aviv2017tbir}
Aviv Eisenschtat and Loir Wolf.
\newblock Linking image and text with 2-way nets.
\newblock In {\em CVPR}, 2017.

\bibitem{tuberlin-dataset}
Mathias Eitz, James Hays, and Marc Alexa.
\newblock How do humans sketch objects?
\newblock {\em ACM TOG}, 2012.

\bibitem{faghri2017tbir}
Fartash Faghri, David~J. Fleet, Jaime~Ryan Kiros, and Sanja Fidler.
\newblock Vse++: Improved visual-semantic embeddings.
\newblock {\em arXiv preprint arXiv:1707.05612}, 2017.

\bibitem{gao2020sketchyCOCO}
Chengying Gao, Qi Liu, Limin Wang, Jianzhuang Liu, and Changqing Zou.
\newblock Sketchycoco: Image generation from freehand scene sketches.
\newblock In {\em CVPR}, 2020.

\bibitem{luis2018textScene}
Llu{\'{\i}}s G{\'{o}}mez, Andr{\'{e}}s Mafla, Mar{\c{c}}al Rusi{\~{n}}ol, and
  Dimosthenis Karatzas.
\newblock Single shot scene text retrieval.
\newblock In {\em ECCV}, 2018.

\bibitem{guo2021scenegraph}
Yuyu Guo, Xuanhan Gao, Liali ad~Wang, Yuxuan Hu, Xing Xu, Xu Lu, Heng~Tao Shen,
  and Jingkuan Song.
\newblock From general to specific: Informative scene graph generation via
  balance adjustment.
\newblock In {\em ICCV}, 2021.

\bibitem{blindcaptioning}
Danna Gurari, Yinan Zhao, Meng Zhang, and Nilavra Bhattacharya.
\newblock Captioning images taken by people who are blind.
\newblock In {\em ECCV}, 2020.

\bibitem{caricatureshop}
Xiaoguang Han, Kangcheng Hou, Dong Du, Yuda Qiu, Yizhou Yu, Kun Zhou, and
  Shugang Cui.
\newblock Caricatureshop: Personalized and photorealistic caricature sketching.
\newblock {\em IEEE TVCG}, 2018.

\bibitem{resnet}
Kaiming He, Xiangyu Zhang, Shaoqing Ren, and Jian Sun.
\newblock Deep residual learning for image recognition.
\newblock In {\em CVPR}, 2016.

\bibitem{hong2021tbir}
Weixiang Hong, Kaixiang Ji, Jiajia Liu, Jian Wang, Jingdong Chen, and Wei Chu.
\newblock Gilbert: Generative vision-language pre-training for image-text
  retrieval.
\newblock In {\em SIGIR}, 2021.

\bibitem{hsu2018disentanglement}
Wei-Ning Hsu and James Glass.
\newblock Disentangling by partitioning: A representation learning framework
  for multimodal sensory data.
\newblock {\em arXiv preprint arXiv:1805.11264}, 2018.

\bibitem{jia2021scalingVL}
Chao Jia, Yinfei Yang, Ye Xia, Yi-Ting Chen, Zarana Parekh, Hieu Pham, Quoc~V.
  Le, Yunhsuan Sung, Zhen Li, and Tom Duerig.
\newblock Scaling up visual and vision-language representation learning with
  noisy text supervision.
\newblock In {\em ICML}, 2021.

\bibitem{deep-visual-semantic}
Andrej Karpathy and Fei-Fei Li.
\newblock Deep visual-semantic alignments for generating image descriptions.
\newblock In {\em CVPR}, 2015.

\bibitem{kim2021ViLT}
Wonjae Kim, Bokyung Son, and Ildoo Kim.
\newblock Vilt: Vision-and-language transformer without convolution or region
  supervision.
\newblock In {\em ICML}, 2021.

\bibitem{glow}
Diederik~P. Kingma and Prafulla Dhariwal.
\newblock Glow: Generative flow with invertible 1x1 convolutions.
\newblock In {\em NeurIPS}, 2018.

\bibitem{gcn}
Thomas~N. Kipf and Max Welling.
\newblock Semi-supervised classification with graph convolutional networks.
\newblock In {\em ICLR}, 2017.

\bibitem{kobayashi2023medical}
Kazuma Kobayashi, Lin Gu, Ryuichiro Hataya, Takaaki Mizuno, Mototaka Miyake,
  Hirokazu Watanabe, Masamichi Takahashi, Yasuyuki Takamizawa, Yukihiro
  Yoshida, Satoshi Nakamura, Nobuji Kouno, Amina Bolatkan, Yusuke Kurose,
  Tatsuya Harada, and Ryuji Hamamoto.
\newblock {Sketch-based Medical Image Retrieval}.
\newblock {\em arXiv preprint arXiv:2303.03633}, 2023.

\bibitem{koley2023picture}
Subhadeep Koley, Ayan~Kumar Bhunia, Aneeshan Sain, Pinaki~Nath Chowdhury, Tao
  Xiang, and Yi-Zhe Song.
\newblock {Picture that Sketch: Photorealistic Image Generation from Abstract
  Sketches}.
\newblock In {\em CVPR}, 2023.

\bibitem{laina2019captioningunsupervised}
Iro Laina, Christian Rupprecht, and Nassir Navab.
\newblock Towards unsupervised image captioning with shared multimodal
  embeddings.
\newblock In {\em ICCV}, 2019.

\bibitem{set-attention}
Juho Lee, Yoonho Lee, Jungtaek Kim, Adam~R. Kosiorek, Seungjin Choi, and
  Yee~Whye Teh.
\newblock Set transformer: A framework for attention-based
  permutation-invariant neural networks.
\newblock In {\em ICML}, 2019.

\bibitem{li2018sketch-r2cnn}
Lei Li, Changqing Zou, Youyi Zheng, Qingkun Su, Honbo Fu, and Chiw-Lan Tai.
\newblock Sketch-r2cnn: An attentive network for vector sketch recognition.
\newblock {\em arXiv preprint arXiv:1811.08170}, 2018.

\bibitem{oscar}
Xiujun Li, Xi Yin, Chunyuan Li, Pengchuan Zhang, Xiaowei Hu, Lei Zhang, Lijuan
  Wang, Houdong Hu, Li Dong, Furu Wei, Yejin Choi, and Jianfeng Gao.
\newblock Oscar: Object-semantics aligned pre-training for vision-language
  tasks.
\newblock In {\em ECCV}, 2020.

\bibitem{lin2004rouge}
Chin-Yew Lin.
\newblock Rouge: A package for automatic evaluation of summaries.
\newblock In {\em Text Summarization Branches Out}, 2004.

\bibitem{lin2020sketchbert}
Hangyu Lin, Yanwei Fu, Yu-Gang Jiang, and Xiangyang Xue.
\newblock Sketch-bert: Learning sketch bidirectional encoder representation
  from transformers by self-supervised learning of sketch gestalt.
\newblock In {\em CVPR}, 2020.

\bibitem{mscoco-dataset}
Tsung-Yi Lin, Michael Maire, Serge~J. Belongie, James Hays, Pietro Perona, Deva
  Ramanan, Piotr Dollár, and C.~Lawrence Zitnick.
\newblock Microsoft coco: common objects in context.
\newblock In {\em ECCV}, 2014.

\bibitem{liu2021radiologycaptioning}
Fenglin Liu, Xian Wu, Shen Ge, Wei Fan, and Yuexian Zou.
\newblock Exploring and distilling posterior and prior knowledge for radiology
  report generation.
\newblock In {\em CVPR}, 2021.

\bibitem{liu2020scenesketcher}
Fang Liu, Changqing Zhou, Xiaoming Deng, Ran Zuo, Yu-Kun Lai, Cuixia Ma,
  Yong-Jin Liu, and Hongan Wang.
\newblock Scenesketcher: Fine-grained image retrieval with scene sketches.
\newblock In {\em ECCV}, 2020.

\bibitem{learn-to-combine}
Kuan Liu, Yanen Li, Ning Xu, and Prem Natarajan.
\newblock Learn to combine modalities in multimodal deep learning.
\newblock {\em arXiv preprint arXiv:1805.11730}, 2018.

\bibitem{liu2018discriminate}
Xihui Liu, Hongsheng Li, Jing Shao, Dapeng Chen, and Xiaogang Wang.
\newblock Show, tell and discriminate: Image captioning by self-retrieval with
  partially labeled data.
\newblock In {\em ECCV}, 2018.

\bibitem{lu2019ViLBERT}
Jiasen Lu, Dhruv Batra, Devi Parikh, and Stefan Lee.
\newblock Vilbert: Pretraining task-agnostic visiolinguistic representations
  for vision-and-language tasks.
\newblock In {\em NeurIPS}, 2019.

\bibitem{mahajan2020lnfmm}
Shweta Mahajan, Iryna Gurevych, and Stefan Roth.
\newblock Latent normalizing flows for many-to-many cross-domain mappings.
\newblock In {\em ICLR}, 2020.

\bibitem{mahajan2020context}
Shweta Mahajan and Stefan Roth.
\newblock Diverse image captioning with context-object split latent spaces.
\newblock In {\em NeurIPS}, 2020.

\bibitem{song2021scaling}
Song Mei, Theodor Misiakiewicz, and Andrea Montanari.
\newblock Learning with invariances in random features and kernel models.
\newblock In {\em CoLT}, 2021.

\bibitem{meng2021mouse}
Zihang Meng, Licheng Yu, Ning Zhang, Tamara Berg, Babak Damavandi, Vikas Singh,
  and Amy Bearman.
\newblock Connecting what to say with where to look by modeling human attention
  traces.
\newblock In {\em CVPR}, 2021.

\bibitem{clipcap}
Ron Mokady, Amir Hertz, and Amit~H. Bermano.
\newblock Clipcap: Clip prefix for image captioning.
\newblock {\em arXiv preprint arXiv:2111.09734}, 2021.

\bibitem{llerena2019cross-modality}
Nils Murrugarra-Llerena and Adriana Kovashka.
\newblock {Cross-Modality Personalization for Retrieval}.
\newblock In {\em CVPR}, 2019.

\bibitem{kaiyue2017cross}
Kaiyue Pang, Yi-Zhe Song, Tao Xiang, and Timothy~M. Hospedales.
\newblock Cross-domain generative learning for fine-grained sketch-based image
  retrieval.
\newblock In {\em BMVC}, 2017.

\bibitem{papineni2002bleu}
Kishore Papineni, Salim Roukos, Todd Ward, and Wei-Jing Zhu.
\newblock Bleu: a method for automatic evaluation of machine translation.
\newblock In {\em ACL}, 2002.

\bibitem{park2017personalised}
Cesc~Chunseong Park, Byeongchang Kim, and Gunhee Kim.
\newblock Attend to you: Personalized image captioning with context sequence
  memory networks.
\newblock In {\em CVPR}, 2017.

\bibitem{park2018personalised}
Cesc~Chunseong Park, Byeongchang Kim, and Gunhee Kim.
\newblock Towards personalized image captioning via multimodal memory networks.
\newblock {\em IEEE TPAMI}, 2018.

\bibitem{plummer2017tbir}
Bryan~A. Plummer, Paige Kordas, M.~Hadi Kaipour, Shuai Zheng, Robinson
  Piramuthu, and Svetlana Lazebnik.
\newblock Conditional image-text embedding networks.
\newblock {\em arXiv preprint arXiv:1711.08389}, 2017.

\bibitem{pont2020mouse}
Jordi Pont-Tuset, Jasper Uijlings, Soravit Changpinyo, Radu Soricut, and
  Vittorio Ferrari.
\newblock Connecting vision and language with localized narratives.
\newblock In {\em ECCV}, 2020.

\bibitem{qi2022segmentation}
Anran Qi, Yulia Gryaditskaya, Tao Xiang, and Yi-Zhe Song.
\newblock One sketch for all: One-shot personalized sketch segmentation.
\newblock {\em TIP}, 2022.

\bibitem{CLIP}
Alec Radford, Jong~Wook Kim, Chris Hallacy, Aditya Ramesh, Gabriel Goh,
  Sandhini Agarwal, Girish Sastry, Amanda Askell, Pamela Mishkin, Jack Clark,
  Gretchen Krueger, and Ilya Sutskever.
\newblock Learning transferable visual models from natural language
  supervision.
\newblock {\em arXiv preprint arXiv:2103.00020}, 2021.

\bibitem{gpt2}
Alec Radford, Jeff Wu, Rewon Child, David Luan, Dario Amodei, and Ilya
  Sutskever.
\newblock Language models are unsupervised multitask learners.
\newblock {\em OpenAI Blog}, 2019.

\bibitem{learning-disentangled-factor}
Scott Reed, Kihyuk Sohn, Yuting Zhang, and Honglak Lee.
\newblock Learning to disentangle factors of variation with manifold
  interaction.
\newblock In {\em ICML}, 2014.

\bibitem{scene-designer}
Leo Sampaio~Ferraz Riberio, Tui Bui, John Collomosse, and Moacir Ponti.
\newblock Scene designer: a unified model for scene search and synthesis from
  sketch.
\newblock In {\em ICCV Workshop}, 2021.

\bibitem{rubenstein2018isotropic}
Paul~K. Rubenstein, Bernhard Schoelkopf, and Ilya Tolstikhin.
\newblock On the latent space of wasserstein auto-encoders.
\newblock {\em arXiv preprint arXiv:1802.03761}, 2018.

\bibitem{sain2023clip}
Aneeshan Sain, Ayan~Kumar Bhunia, Pinaki~Nath Chowdhury, Aneeshan Sain,
  Subhadeep Koley, Tao Xiang, and Yi-Zhe Song.
\newblock {CLIP for All Things Zero-Shot Sketch-Based Image Retrieval,
  Fine-Grained or Not}.
\newblock In {\em CVPR}, 2023.

\bibitem{sain2023exploiting}
Aneeshan Sain, Ayan~Kumar Bhunia, Subhadeep Koley, Pinaki~Nath Chowdhury,
  Soumitri Chattopadhyay, Tao Xiang, and Yi-Zhe Song.
\newblock {Exploiting Unlabelled Photos for Stronger Fine-Grained SBIR}.
\newblock In {\em CVPR}, 2023.

\bibitem{sain2022sketch3t}
Aneeshan Sain, Ayan~Kumar Bhunia, Vaishnav Potlapalli, Pinaki~Nath Chowdhury,
  Tao Xiang, and Yi-Zhe Song.
\newblock Sketch3t: Test-time training for zero-shot sbir.
\newblock In {\em CVPR}, 2022.

\bibitem{sain2020crossmodal}
Aneeshan Sain, Ayan~Kumar Bhunia, Yongxin Yang, Tao Xiang, and Yi-Zhe Song.
\newblock Cross-modal hierarchical modelling for fine-grained sketch based
  image retrieval.
\newblock In {\em BMVC}, 2020.

\bibitem{styleMeUp}
Aneeshan Sain, Ayan~Kumar Bhunia, Yongxin Yang, Tao Xiang, and Yi-Zhe Song.
\newblock Stylemeup: Towards style-agnostic sketch-based image retrieval.
\newblock In {\em CVPR}, 2021.

\bibitem{sangkloy2022textSketch}
Patsorn Sangkloy, Wittawat Jitkrittum, Diyi Yang, and James Hays.
\newblock A sketch is worth a thousand words: Image retrieval with text and
  sketch.
\newblock In {\em ECCV}, 2022.

\bibitem{transformer}
Vaswani~A. Shazeer, N. Parmar, N. Uszkoreit, J. Jones, A.~N. Gomez, L. Kaiser,
  and I. Polosukhin.
\newblock Attention is all you need.
\newblock In {\em NeurIPS}, 2017.

\bibitem{shuster2019personality}
Kurt Shuster, Samuel Humeau, Hexiang Hu, Antonie Bordes, and Jason Weston.
\newblock Engaging image captioning via personality.
\newblock In {\em CVPR}, 2019.

\bibitem{vgg-16}
Karen Simonyan and Andrew Zisserman.
\newblock Very deep convolutional networks for large-scale image recognition.
\newblock In {\em ICLR}, 2015.

\bibitem{song2017textSketch}
Jifei Song, Yi-Zhe Song, Tao Xiang, and Timothy Hospedales.
\newblock Fine-grained image retrieval: the text/sketch input dilemma.
\newblock In {\em BMVC}, 2017.

\bibitem{deep-spatial-semantic}
Jifei Song, Qian Yu, Yi-Zhe Song, Tao Xiang, and Timothy~M Hospedales.
\newblock Deep spatial-semantic attention for fine-grained sketch-based image
  retrieval.
\newblock In {\em ICCV}, 2017.

\bibitem{spurr2018crossmodal}
Adrian Spurr, Jie Song, Seonwook Park, and Otmar Hilliges.
\newblock Cross-modal deep variational hand pose estimation.
\newblock In {\em CVPR}, 2018.

\bibitem{stefanini2021captioningsurvey}
Matteo Stefanini, Marcella Cornia, Lorenzo Baraldi, Silvia Cascianelli,
  Giuseppe Fiameni, and Rita Cucchiara.
\newblock From show to tell: A survey on deep learning-based image captioning.
\newblock {\em arXiv preprint arXiv:2208.04254}, 2021.

\bibitem{torontofacedataset}
Joshua Susskind, Adam Anderson, and Geoffrey Hinton.
\newblock The toronto face dataset.
\newblock Technical report, Toronto University, 2010.

\bibitem{suzuki2017disentanglement}
Masahiro Suzuki, Kotaro Nakayama, and Yutaka Matsuo.
\newblock Joint multimodal learning with deep generative models.
\newblock In {\em ICLR Workshop}, 2017.

\bibitem{touvron2021kd}
Hugo Touvron, Matthieu Cord, Matthijs Douze, Francisco Massa, Alexandre
  Sablayrolles, and Hervé Jégou.
\newblock Training data-efficient image transformers \& distillation through
  attention.
\newblock In {\em ICML}, 2021.

\bibitem{tripathi2020object}
Aditay Tripathi, Rajath~R. Dani, Anand Mishra, and Anirban Chakraborty.
\newblock Sketch-guided object localization in natural images.
\newblock In {\em ECCV}, 2020.

\bibitem{learning-factorised-representations}
Yao-Hung~Hubert Tsai, Paul~Pu Liang, Amir Zadeh, Louis-Philippe Morency, and
  Ruslan Salakhutdinov.
\newblock Learning factorized multimodal representations.
\newblock In {\em ICLR}, 2019.

\bibitem{cpc2018}
Aaron van~den Oord, Yazhe Li, and Oriol Vinyals.
\newblock Representation learning with contrastive predictive coding.
\newblock {\em arXiv preprint arXiv:1807.03748}, 2018.

\bibitem{vedantam2015cider}
Ramakrishna Vedantam, C.~Lawrence Zitnick, and Devi Parikh.
\newblock Cider: Consensus-based image description evaluation.
\newblock In {\em CVPR}, 2015.

\bibitem{veit2018separating}
Andreas Veit, Maximilian Nickel, Serge Belongie, and Laurens van~der Maaten.
\newblock {Separating Self-Expression and Visual Content in Hashtag
  Supervision}.
\newblock In {\em CVPR}, 2018.

\bibitem{wang2017agcvae}
Liwei Wang, Alexander~G. Schwing, and Svetlana Lazebnik.
\newblock Diverse and accurate image description using a variational
  auto-encoder with an additive gaussian encoding space.
\newblock In {\em NeurIPS}, 2017.

\bibitem{wang2022medical}
Xi Wang, Kathleen Ang, and Faramarz Samavati.
\newblock Sketch-based editing and deformation of cardiac image segmentation,
  2022.

\bibitem{camp}
Zihao Wang, Xihui Liu, Hongsheng Li, Lu Sheng, Junjie Yan, Xiaogang Wang, and
  Jing Shao.
\newblock Camp: Cross-modal adaptive message passing for text-image retrieval.
\newblock In {\em ICCV}, 2019.

\bibitem{alt-text}
Shaomei Wu, Jeffrey Wieland, Omid Farivar, and Julie Schiller.
\newblock Automatic alt-text: Computer-generated image descriptions for blind
  users on a social network service.
\newblock In {\em CSCW}, 2017.

\bibitem{xing2015autocomplete}
Jun Xing, Li-Yi Wei, Takaaki Shiratori, and Koji Yatani.
\newblock Autocomplete hand-drawn animations.
\newblock {\em ACM TOG}, 2015.

\bibitem{xu2015show}
Kelvin Xu, Jimmy Ba, Ryan Kiros, Kyunghyun Cho, Aaron Courville, Ruslan
  Salakhutdinov, Richard Zemel, and Yoshua Bengio.
\newblock Show, attend and tell: Neural image caption generation with visual
  attention.
\newblock In {\em ICML}, 2015.

\bibitem{xu2022survey}
Peng Xu, Timothy~M. Hospedales, Qiyue Yin, Yi-Zhe Song, Tao Xiang, and Liang
  Wang.
\newblock Deep learning for free-hand sketch: A survey.
\newblock {\em IEEE TPAMI}, 2022.

\bibitem{xu2022MML}
Peng Xu, Xiatian Zhu, and David~A. Clifton.
\newblock Multimodal learning with transformers: A survey.
\newblock {\em arXiv preprint arXiv:2206.06488}, 2022.

\bibitem{xue2021intermodality}
Hongwei Xue, Yupan Huang, Bei Liu, Houwen Peng, Jianlong Fu, Houqiang Li, and
  Jiebo Luo.
\newblock Probing inter-modality: Visual parsing with self-attention for
  vision-language pre-training.
\newblock In {\em NeurIPS}, 2021.

\bibitem{yang2020surgery}
Shuai Yang, Zhangyang Wang, Jiaying Liu, and Zongming Guo.
\newblock Deep plastic surgery: Robust and controllable image editing with
  human-drawn sketches.
\newblock In {\em ECCV}, 2020.

\bibitem{yelamarthi2018sketch}
Sasi~Kiran Yelamarthi, Shiva~Krishna Reddy, Ashish Mishra, and Anurag Mittal.
\newblock A zero-shot framework for sketch based image retrieval.
\newblock In {\em ECCV}, 2018.

\bibitem{yu2016shoe}
Qian Yu, Feng Liu, Yi-Zhe Song, Tao Xiang, Timothy~M. Hospedales, and
  Chen~Change Loy.
\newblock Sketch me that shoe.
\newblock In {\em CVPR}, 2016.

\bibitem{sketch-a-net}
Qian Yu, Yongxin Yang, Feng Liu, Yi-Zhe Song, Tao Xiang, and Timothy~M
  Hospedales.
\newblock Sketch-a-net: A deep neural network that beats humans.
\newblock {\em IJCV}, 2017.

\bibitem{self-atten-gan}
Han Zhang, Ian~J. Goodfellow, Dimitris~N. Metaxas, and Augustus Odena.
\newblock Self-attention generative adversarial networks.
\newblock In {\em ICML}, 2019.

\bibitem{stackgan}
Han Zhang, Tao Xu, Hongsheng Li, Shaoting Zhang, Xiaogang Wang, Xialei Huang,
  and Dimitris~N. Metaxas.
\newblock Stackgan++: Realistic image synthesis with stacked generative
  adversarial networks.
\newblock In {\em IEEE TPAMI}, 2019.

\bibitem{zhang2020personalised}
Wei Zhang, Yue Ying, Pang Lu, and Hongyuan Zha.
\newblock Learning long- and short-term user literal-preference with multimodal
  hierarchical transformer network for personalized image caption.
\newblock In {\em AAAI}, 2020.

\bibitem{zhang2022captioning}
Youyan Zhang, Jiuniu Wang, Hao Wu, and Wenjia Xu.
\newblock Distinctive image captioning via clip guided group optimization.
\newblock {\em arXiv preprint arXiv:2208.04254}, 2022.

\bibitem{places-dataset}
Bolei Zhou, Agata Lapedriza, Aditya Khosla, Aude Olivia, and Antonio Torralba.
\newblock Places: A 10 million image database for scene recognition.
\newblock {\em IEEE TPAMI}, 2017.

\bibitem{zhou2016semantic}
Bolei Zhou, Hang Zhao, Xavier Puig, Sanja Fidler, Adela Barriuso, and Antonio
  Torralba.
\newblock Semantic understanding of scenes through the ade20k dataset.
\newblock {\em IJCV}, 2019.

\bibitem{zou2018sketchyscene}
Changqing Zou, Qian Yu, Ruofei Du, Haoran Mo, Yi-Zhe Song, Tao Xiang, Chengying
  Gao, Baoquan Chen, and Hao Zhang.
\newblock Sketchyscene: Richly-annotated scene sketches.
\newblock In {\em ECCV}, 2018.

\end{thebibliography}
}

\end{document}